\pdfoutput=1

\documentclass[11pt]{article}

\usepackage[final]{acl}

\usepackage{times}
\usepackage{latexsym}

\usepackage[T1]{fontenc}

\usepackage[utf8]{inputenc}

\usepackage{microtype}

\usepackage{inconsolata}

\usepackage{graphicx}

\usepackage{color}

\usepackage{footnote}
\usepackage{tabularx}
\usepackage{booktabs}
\usepackage{adjustbox}
\usepackage{multirow}
\usepackage{makecell}

\usepackage{amsmath}
\usepackage{amssymb}
\usepackage{dsfont}
\usepackage{algorithmicx}
\usepackage{algpseudocode}
\usepackage[linesnumbered,ruled,vlined]{algorithm2e}
\usepackage{lipsum}  
\usepackage{float}
\usepackage{caption}
\usepackage{algorithm2e} 
\usepackage{bbm}
\usepackage{xcolor}
\usepackage{xspace}
\usepackage{hyperref}
\usepackage[defaultcolor=magenta]{changes}

\newcommand{\modelname}{R2-KG\xspace}
\newcommand{\sag}{\textit{Supervisor}\xspace} 
\newcommand{\oag}{\textit{Operator}\xspace} 

\DeclareMathOperator*{\argmax}{arg\,max}

\title{\modelname: General-Purpose Dual-Agent Framework for\\ Reliable Reasoning on Knowledge Graphs}

\author{
  Sumin Jo\thanks{Equal contribution.}, Junseong Choi\footnotemark[1], Jiho Kim, Edward Choi \\
  KAIST \\
  \texttt{\{ekrxjwh2009, quasar0311, jiho.kim, edwardchoi\}@kaist.ac.kr}
}

\begin{document}
\maketitle
\begin{abstract}
Recent studies have combined Large Language Models (LLMs) with Knowledge Graphs (KGs) to enhance reasoning, improving inference accuracy without additional training while mitigating hallucination. 
However, existing frameworks still suffer two practical drawbacks: they must be re-tuned whenever the KG or reasoning task changes, and they depend on a single, high-capacity LLM for reliable (\textit{i.e.,} trustworthy) reasoning.
To address this, we introduce \textit{\modelname}, a plug-and-play, dual-agent framework that separates reasoning into two roles: an \oag that gathers evidence and a \sag that makes final judgments.
This design is cost-efficient for LLM inference while still maintaining strong reasoning accuracy.
Additionally, \modelname employs an \textit{Abstention mechanism}, generating answers only when sufficient evidence is collected from KG, which significantly enhances reliability.
Experiments across five diverse benchmarks show that \modelname consistently outperforms baselines in both accuracy and reliability, regardless of the inherent capability of LLMs used as the \oag. Further experiments reveal that the single-agent version of \modelname, equipped with a strict self-consistency strategy, achieves significantly higher-than-baseline reliability with reduced inference cost but increased abstention rate in complex KGs.
Our findings establish \modelname as a flexible and cost-effective solution for KG-based reasoning, reducing reliance on high-capacity LLMs while ensuring trustworthy inference. 
\footnote{The code is available at \url{https://github.com/ekrxjwh2009/R2-KG.git}.}

\end{abstract}

\section{Introduction}
Recent studies have increasingly integrated Large Language Models (LLMs) with Knowledge Graphs (KGs) to perform knowledge-grounded reasoning \cite{generateOnGraph, kim2024causalreasoninglargelanguage, gao2024twostagegenerativequestionanswering, luo2024graphconstrainedreasoningfaithfulreasoning, ma2024debategraphflexiblereliable, reasoningefficientknowledgepathsknowledge}.
This approach maximizes reasoning performance by combining the domain-specific knowledge of KGs with the strong reasoning abilities of LLMs \cite{KG_survey1, zhu2024llmsknowledgegraphconstruction}.

An agent-based LLM framework treats the LLM itself as an agent that selects actions in KG and then generates the final answer~\cite{ToG, ToG2.0, kg-agent2024}. Existing agent-based LLM frameworks \emph{claim} to be task- and KG-agnostic, yet in practice, they require
non-trivial manual effort whenever either the knowledge graph changes (\textit{e.g.}, DBpedia \cite{lehmann2015dbpedia} $\rightarrow$ Freebase \cite{10.1145/1376616.1376746}) or new reasoning task is introduced (\textit{e.g.}, question answering $\rightarrow$ fact verification) \cite{kim-etal-2023-kg,ToG,ToG2.0}. For example, one of the most prominent frameworks—\textit{Think-on-Graph} (ToG)~\cite{ToG} reaches its reported score only after users hand-tune exploration hyperparameters: \textit{depth} and \textit{width}. Also, moving to a temporal KG demands direct modification of the algorithm to inject time-aware pruning.
Such hidden costs fall entirely on practitioners and undermine the promise of true generalizability.

\begin{figure}[t]
    \centering
    \includegraphics[width=\linewidth]{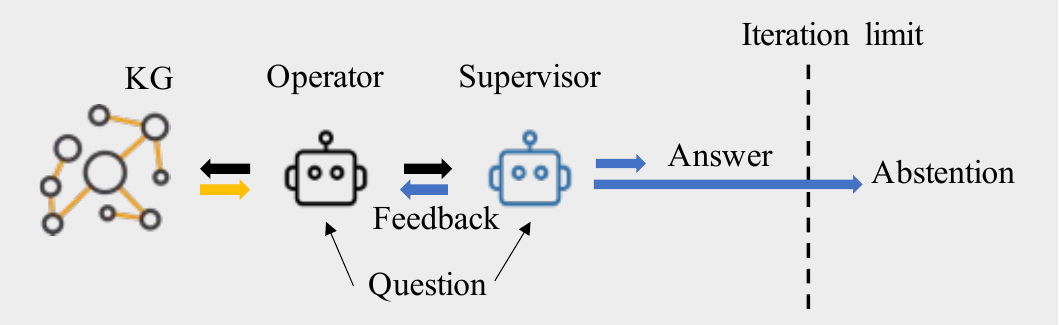}
    \caption{\modelname: The two agents provide an `Answer' only when they are confident enough to do so. If multiple attempts at exploration fail to gather sufficient information, it determines that it does not know and abstains from answering.}
    \label{fig:enter-label}
\end{figure}

Moreover, existing single-agent frameworks rely on one LLM to handle both KG exploration and answer generation, so their overall robustness is tightly coupled to that model’s capacity. Low-capacity LLMs are more prone to early KG retrieval errors, and lack any built-in mechanism to detect or correct those mistakes.  
As the same agent prunes paths while exploring the KG, it cannot revisit discarded branches, resulting in an \textbf{irreversible search}.  
Once the forward path diverges, the system can be trapped in an incorrect subgraph without any means of recovery \cite{ToG2.0,pathsovergraph,ToG,kg-agent2024}.

\textbf{\modelname eliminates these issues.} We decouple evidence collection and answer validation into two collaborating agents: the \oag that explores the KG and logs every \(<\!{\tt entity}, {\tt relation}\!>\) decision in a persistent \textit{chat log}, and the \sag that (i) audits the current evidence or (ii) may issue a \textit{feedback} command to return to an earlier hop or explore an unexplored path.
Through this iterative collaboration, if the framework exceeds a fixed iteration limit, it automatically abstains from answering (\textit{i.e., Abstention mechanism}).
By answering only when the evidence is sufficient—and otherwise opting not to answer—we ensure reliability. 
Thanks to the \oag's parallel exploration strategy, \textbf{\modelname} can keep \emph{multiple} candidate paths alive and expand them concurrently—an ability not supported by such approaches as ToG \cite{ToG} or KG-GPT \cite{kim-etal-2023-kg}.
Furthermore, the reasoning logic of \modelname is frozen, thus porting \modelname to a new KG or task requires only swapping the in-context examples in the prompt.
Compared to SOTA single‑agent frameworks, our work contributes:

\textbf{(1) Dual-Agent Separation for Accuracy and Cost Efficiency}—The low-capacity \oag handles KG exploration, while the high-capacity \sag provides path-level feedback and generates the final answer. The \oag can explore \textit{multiple} candidate paths in parallel, while the \sag provides feedback that steers the \oag toward more promising branches; leveraging the \textit{chat log}, the \oag can roll back to any earlier hop and re‑route when necessary. Even when both agents run on low‑capacity LLMs, \modelname surpasses the best‑reported performance of SOTA baselines, underscoring the strength of the architecture itself.
This division increases overall accuracy while reducing the overall LLM cost.

\textbf{(2) KG- and Task-Agnostic Plug-and-Play Deployment}—Porting \modelname to a new KG or reasoning task requires only swapping entity or relation names in the in-context examples, without any substantial hyperparameter tuning or algorithm edits. We evaluated on five diverse benchmarks—covering fact verification \cite{kim-etal-2023-factkg}, single-label QA, multi-label QA \cite{webqsp,metaQA,talmor-berant-2018-web}, and temporal QA \cite{cronQA}—\modelname surpasses strong baselines, attaining a 100\% hit rate on MetaQA \cite{metaQA} and up to +87.8\% micro-F1 over the previous SOTA.

\textbf{(3) Reliability Through the Abstention Mechanism}—The \sag defers answering until evidence is sufficient; otherwise \modelname returns \textbf{Abstain}. As a result, \modelname offers high F1 and hit rates when it does answer, and refrains when it cannot ground a claim—maintaining user trust even when driven by low-capacity \oag models.

\textbf{(4) Single-Agent Version with Strict Self-Consistency for Further Cost Savings}—We propose an even more cost-efficient method that eliminates \sag (\textit{i.e.}, single-agent version of \modelname combined with strict self-consistency strategy \cite{wang2023selfconsistencyimproveschainthought}). Here, the low-capacity \oag alone conducts reasoning, but ensures high reliability by requiring unanimous agreement across multiple trials before producing a result.
This approach further reduces inference cost significantly, but comes with a trade-off of increased abstention rate, particularly in complex KGs with temporal information.

\begin{figure*}[ht]
    \centering
    \includegraphics[width=\textwidth]{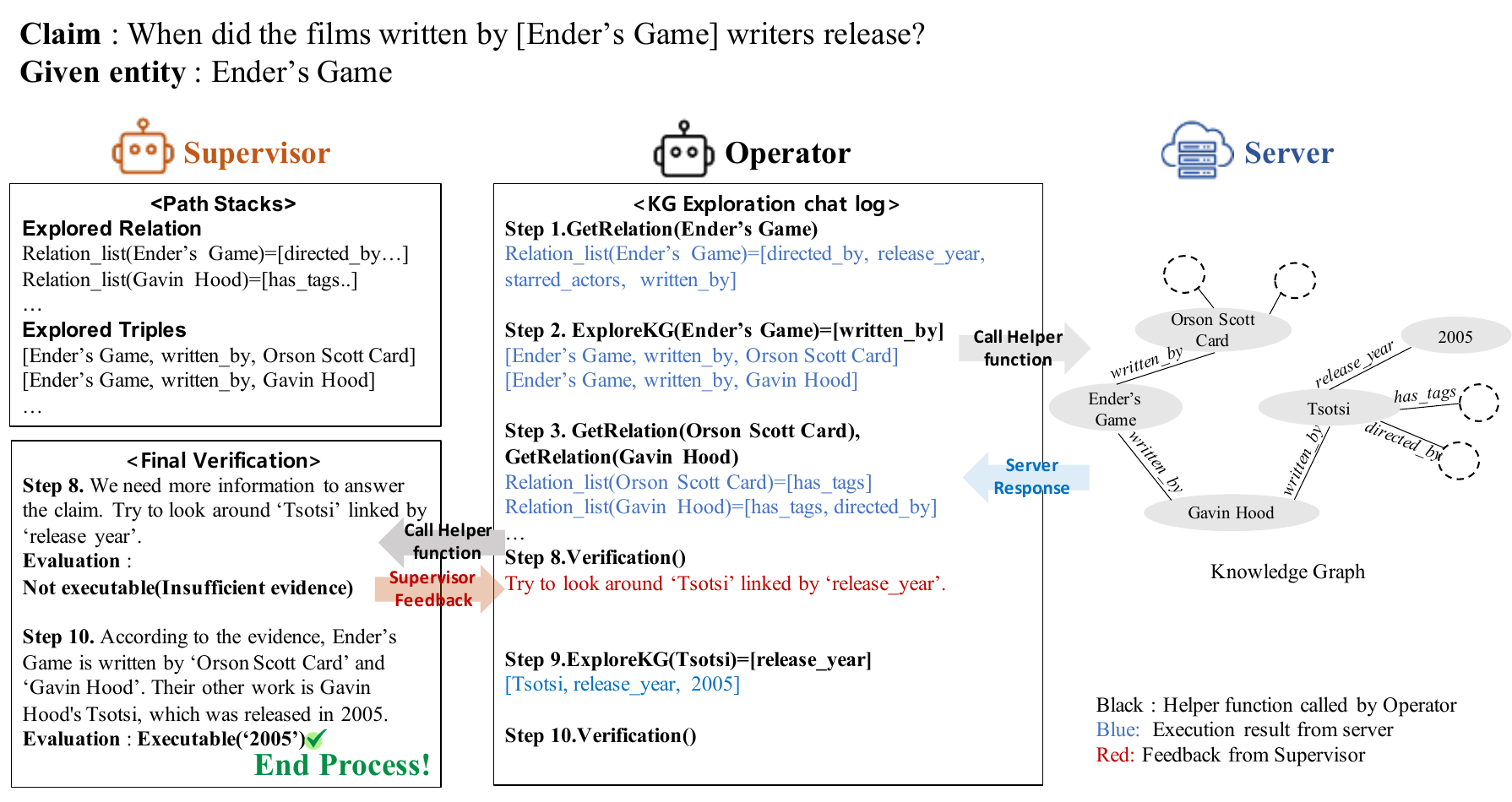}
    \caption{\modelname solves multi-hop query through an iterative dialogue between a low-capacity \oag and a high-capacity \sag. The \oag gathers triples via \textit{GetRelation()} and \textit{ExploreKG()} calls, and all of the explored relations ($R_k$) and explored triples ($G_k$) are stacked in the \sag's \textit{Path Stacks} at every step $k < T (iteration\;limit)$. According to the \textit{Path Stacks}, if evidence is lacking for the verification, the \sag sends feedback to the \oag to pursue alternative paths or roll back to an earlier hop.}
    \label{fig:enter-label}
\end{figure*}

\section{Related Works}

\subsection{KG-Based Reasoning with LLM}
\label{sec:related_kg_based}
Research on KG-based reasoning tasks can be broadly categorized into three approaches: embedding-based, semantic parsing-based, and retrieval-augmented \cite{lan2022complexknowledgebasequestion, ji-etal-2024-retrieval, mavromatis2024gnnraggraphneuralretrieval}. First, the embedding-based method projects the entities and relations of a KG into an embedding space \cite{saxena-etal-2020-improving}. This approach effectively captures complex relationships and multi-hop connections through vector operations.

Second, the semantic parsing-based method converts the task into a symbolic logic form (\textit{e.g.}, a SPARQL query \cite{perez2009semantics}) and executes it on the KG to derive the final answer \cite{Sun_Zhang_Cheng_Qu_2020, park2021knowledge, ye-etal-2022-rng, gu-su-2022-arcaneqa, yu2023decafjointdecodinganswers}. This approach has the advantage of handling complex queries, such as multi-hop reasoning, through intuitive queries that can be directly applied to the KG.

Third, the retrieval-augmented method extracts relevant subgraphs from the KG to infer the answers. Recent studies explored using LLMs for both retrieval and reasoning without additional training \cite{kim-etal-2023-kg, wang2023knowledgedrivencotexploringfaithful, structgpt, li-etal-2023-shot, ToG, ToG2.0}. KG-GPT \cite{kim-etal-2023-kg} proposed a three-stage framework: Sentence Segmentation, Graph Retrieval, and Inference. ToG \cite{ToG} and ToG-2.0 \cite{ToG2.0} introduced frameworks that conduct reasoning by pruning relations and entities during KG exploration. While these LLM-based methods enhance the performance of KG-based reasoning, they struggle to adapt to new KG structures or tasks. 
Also, these frameworks can explore the KG only up to the fixed hyperparameters (\textit{e.g., }depth, width, top-k), and because they do not retain the full history of visited triples, they cannot return to earlier paths. As a result, potentially relevant branches can be missed. To overcome these limitations, we introduce \modelname, a truly generalizable framework that enables more accurate and efficient KG exploration.

\subsection{Enhancing Model Reliability via Abstention Mechanism}
To mitigate LLM hallucination, the \textit{abstention mechanism} has been adopted as a strategy to enhance reliability \cite{wen2024knowlimitssurveyabstention}. This mechanism allows the model to refrain from answering when the input query is ambiguous \cite{asai-choi-2021-challenges, cole-etal-2023-selectively}, goes against human values \cite{kirk-etal-2023-past}, or exceeds the model's knowledge scope \cite{feng-etal-2024-dont}.  
The \textit{abstention mechanism} has been actively explored in LLM-based question-answering tasks, particularly for long-document processing QA \cite{buchmann-etal-2024-attribute} and uncertainty estimation \cite{amayuelas-etal-2024-knowledge, wen-etal-2024-characterizing, yang2024alignmenthonesty, tomani2024uncertaintybasedabstentionllmsimproves}, demonstrating notable improvements in reliability.
However, its application in KG-based reasoning remains largely unexplored.
We introduce \textit{Reliable KG-Based Reasoning Task}, the first approach to integrate the \textit{abstention mechanism} into KG-based reasoning.

\section{Reliable KG-Based Reasoning Task}
\subsection{Task Definition}
In this study, we propose the \textit{Reliable KG-Based Reasoning Task} for the first time.
This task serves as a benchmark for measuring reliability in KG-based reasoning, particularly in domains where trustworthy answers are critical, such as industrial applications and fact verification that utilize KGs.
By evaluating reliability, this enables the selection of an appropriate framework based on the specific context. 
Unlike existing KG-based reasoning tasks that focus on generating a definitive answer \textit{a} (\textit{e.g.}, True / False in fact verification or a direct response in QA) for a given query \textit{q} (\textit{e.g.}, a query in fact verification or a question in QA), our task introduces the option to \textit{abstain} when uncertainty arises.
This allows the system to either withhold a response when sufficient evidence cannot be retrieved from the KG or avoid providing an unreliable answer based on ambiguous evidence.

\subsection{Metrics}
To evaluate the KG-based reasoning task incorporating the \textit{abstention mechanism}, we measure four key metrics:

\textbf{Coverage}: The fraction of samples for which a final answer is generated (\textit{i.e.}, the ratio of non-abstained samples). 
\begin{equation}
\small
   \text{Coverage} = \frac{|\mathcal{S}|}{|N|}
\nonumber
\end{equation}
where \(\mathcal{S}\) denotes the set of non-abstained samples, and \(N\) represents the set of all samples, including abstained and non-abstained cases.
    

\textbf{Micro F1 Score}: Computed on $\mathcal{S}$ in multi-label tasks using \(TP_i, FP_i, FN_i\), which represent the True Positives, False Positives, and False Negatives for each sample \(i\), respectively.
    \begin{equation}
    \small
    \begin{aligned}
        \text{Micro F1} = \frac{2 \times \text{Total Precision} \times \text{Total Recall}}{\text{Total Precision} + \text{Total Recall}}\quad\quad\quad\\[5pt]
        \substack{\text{Total} \\ \text{Precision}} = \frac{\sum_{i \in \mathcal{S}}  TP_{i}}{\sum_{i \in \mathcal{S}}( TP_{i} + FP_{i})}, 
        \substack{\text{Total} \\ \text{Recall}} = \frac{\sum_{i \in \mathcal{S}}  TP_{i}}{\sum_{i \in \mathcal{S}}(  TP_{i} +   FN_{i})}
    \end{aligned}
    \nonumber
\end{equation}

\textbf{Samplewise F1 Score}: Calculated on $\mathcal{S}$ in multi-label tasks by computing F1 score for each sample and averaging over \(\mathcal{S}\).
    \begin{equation}
    \small
    \begin{aligned}
        \text{Samplewise F1} = \frac{1}{|\mathcal{S}|} \sum_{i \in \mathcal{S}} 
        \frac{2 \times \text{Precision}_i \times \text{Recall}_i}
        {\text{Precision}_i + \text{Recall}_i} \\
        \text{Precision}_i = \frac{TP_{i}}{TP_{i} + FP_{i}}, \text{Recall}_i = \frac{TP_{i}}{TP_{i} + FN_{i}}
    \end{aligned}
    \nonumber
\end{equation}

\textbf{Hit Rate}: Applicable to both single-label and multi-label tasks. It is counted if any predicted label matches a ground-truth label. Note that the hit rate is the accuracy in binary tasks.
\begin{equation}
\small
\text{Hit rate} = \frac{1}{|\mathcal{S}|} \sum_{i \in \mathcal{S}} \mathds{1}(\hat{y}_i \in Y_i)
\label{eq:hit_rate}
\nonumber
\end{equation}
where \(\mathds{1}(\cdot)\) is the indicator function, \(\hat{y}_i\) is one of the framework’s predicted label for sample \(i\) and \( Y_i\) is the set of ground truth labels for sample \(i\).

\section{Method}
Our \modelname consists of three components: An \oag, which explores the KG via helper functions; a \textit{Server}, which provides requested function output; and a \sag, which offers feedback or generates the final answer.
Within an iteration limit \textit{T}, the three components iteratively interacts, gathering triples \(G_t\) or relations \(R_t\), at each step \(t\). The \sag outputs the final answer once sufficient evidence is collected. If no answer is produced within \textit{T}, the system returns an \textit{Abstention}, indicating insufficient understanding of the query.

\subsection{Operator}

By leveraging helper functions (described below), the system retrieves relevant subgraphs from the KG. When the \oag requests a function call, the \textit{Server} responds, and their interactions are accumulated in \textit{chat log} at each step \textit{t} for future reference. 

For multi-hop reasoning, \modelname iteratively expands the subgraphs by accumulating relevant triples.
Given a query where entity $e_{0}$ and $e_{n}$ are connected through \(n\)-hops, the intermediate entities are unknown.
At an arbitrary step \(k\), the \oag maintains $E^{(t=k)}_{seen} =\{ e_0, \dots,e_{m-1}, e_m \}$, which is the set of entities explored up to the previous step, where \(E_{seen}^{(t=0)}=\left\{e^{0}\right\}\). Each \(e_i \in E_{seen}\) is associated with relations $R(e_i) = \{ r_{i(1)}, r_{i(2)}, \dots, r_{i(n)} \}$.
In the next step, \oag selects a relevant \(e^* \in E_{seen} \) and one or more relevant relations $R^{*} \subseteq R(e^{*})$, retrieves the corresponding tail entities, and get a new triple set: $\{(e^*, r^{*}, e_{m+1}) \mid r^{*} \in R^{*} \}$. This process continues until \(e_{m+1}\) matches \(e_{n}\).

By structuring reasoning in this way, \modelname ensures that each step builds upon \textit{chat log}, improving both exploration efficiency and reasoning accuracy. The \oag can, at each step-$t$, invoke \textit{multiple} following helper functions in parallel enabling simultaneous exploration of several graph branches and accelerating KG search.

\textbf{GetRelation}(\( e^{*} \)): The \textit{Server} returns all relations \(R(e^{*}) \) connected to $e^*$ in the KG as follows:\\[3pt]
\begin{equation}
\small
\begin{aligned}
    e^{*} &= \argmax_{e \in E_{seen}} EntScore(e, q) \\ 
    R(e^{*}) &= \left\{ r_{i} \mid (e^{*},r_{i}, e_{j})\in KG, \forall e_{j}   \right\} \nonumber
\end{aligned}
\label{eq:label}    
\end{equation}

 The \oag selects \(e^{*}\) that is most relevant to \textit{q} among \(E_{seen}\) using \(EntScore(e, q)\), which is a function that evaluates the relevance between \(e\) and \textit{q}.
 Note that \(EntScore(\cdot)\) is based not on an explicit implementation but on the inherent language understanding of the \oag.

\textbf{ExploreKG}\( (e^{*}, R^{*}(e^{*})) \): The \textit{Server} returns \( G(e^{*}, R^{*}(e^*))\), a set of all triples such that \(e^{*} \in E_{seen}\) is connected to a tail entity \(e_j\) via the relation \(r_i \in R^*(e^*)\). Note that \( R^*(e^{*})\) is a subset of \( R(e^{*})\), which is returned by \textit{GetRelation()} chosen by \textit{RelScore()} as below:
\begin{equation}
\small
\begin{array}{c}
    R^*(e^{*}) = \left\{ r \mid r \in R(e^{*}), \; RelScore(r, q) > \text{threshold} \right\} 
    \\[5pt]
    G(e^{*}, R^{*}(e^*)) = \left\{ (e^{*}, r_{i}, e_{j}) \mid r_i \in R^{*}(e^*), \; e_{j} \in KG \right\} \nonumber
\end{array}
\label{eq:your_label}
\end{equation}

\noindent\(RelScore(r, q)\) evaluates the relevance between \(r\) and \(q\) based on the inherent language understanding of the \oag.
Along with the threshold, it is implicitly applied during the \oag's linguistic reasoning process to select several relations relevant to \(q\). 

\textbf{Verification}(\(G_{k}\), \(R_{k}\)): If the collected evidence is deemed sufficient, \oag invokes the \sag. The \oag provides the explored triples \(G_{k}\) and explored relations \(R_{k}\) gathered up to the current step \(k (<T)\) to the \sag. If the \sag gives back an answer, the process terminates; otherwise, if feedback is given, the next iteration continues.\\
\begin{equation}
\small
\begin{aligned}
R_{\text{k}} = \bigcup\limits_{t=1}^{k} R_{t}(e^{*}), \quad
G_{\text{k}} = \bigcup\limits_{t=1}^{k} G_{t}(e^{*}, R^{*}(e^{*}))
\nonumber
\label{eq:label}
\end{aligned}
\end{equation}

\subsection{Supervisor}
The \sag performs its role only when the \oag invokes \textit{Verification(\(G_{k}\), \(R_{k}\))}. Upon invocation, the \sag receives the $G_k$ and $R_k$ and returns one of two possible outcomes to the \oag:\\
\textbf{1) Sufficient Evidence (answer):} If sufficient information is available, the \sag generates a prediction and returns it to the \oag. The final reasoning path\footnote{You can find the example of final reasoning path of \sag from Appendix~\ref{sec:finalgoldpath}} optimized for answer generation is constructed by the \sag based on its judgment, using \(G_k\).\\
\textbf{2) Insufficient Evidence:} If the evidence is lacking, based on $G_k$, $R_k$, and $q$, the \sag suggests new or previously pruned \emph{entity–relation} pairs, enabling the \oag to roll back to any earlier hop or branch into unseen entities before resuming the search\footnote{You can find the example of \sag's feedback for \oag in Appendix~\ref{sec:qualitative_analysis}}.

\subsection{Configurable Iteration Limit}


During KG exploration, \modelname requires at least two iterations to traverse a single hop---first using \textit{GetRelation(\(\cdot\))}, then \textit{ExploreKG(\(\cdot\))}. Therefore, if a query requires \(H\) hops, we recommend setting the iteration limit \(T \geq 2H\). However, since \modelname can issue multiple helper function calls in parallel within an iteration and flexibly reuse partial evidence, it can complete multi-hop reasoning with fewer than \(2H\) steps.

Unlike prior methods where hyperparameters (\textit{e.g.}, depth in ToG, top-\(k\) in KG-GPT) directly constrain the discoverable reasoning paths, \(T\) in \modelname is designed as a reliability knob. A lower \(T\) favors high-precision decisions by limiting uncertain inferences, while a higher \(T\) enhances coverage through broader evidence exploration.

\section{Experiments}

\subsection{Datasets}

To demonstrate that \modelname is a plug-and-play approach independent of task and KG variation, we use five challenging benchmarks with diverse query difficulty, KG structures, and task formats. Table~\ref{tab:data-statistics} shows the features and statistics of the dataset we used.
WebQSP \cite{webqsp} is a semantic parsing QA dataset, and CWQ \cite{talmor-berant-2018-web} builds on it with more complex questions involving compositional and comparative reasoning.
MetaQA \cite{metaQA} dataset has 1-hop, 2-hop, and 3-hop questions, we focus on most challenging 3-hop task\footnote{MetaQA 1-hop and 2-hop tasks are covered in Appendix~\ref{sec:metaqa_12hop}}.  
CRONQUESTIONS \cite{cronQA} is a temporal reasoning benchmark, we used three question types (\textit{i.e.}, simple time, simple entity, time join), excluding others due to missing labels (details in Appendix~\ref{sec:cronq_excluded}).  
FactKG \cite{kim-etal-2023-factkg} contains the most structurally complex multi-hop queries among publicly released benchmarks to date with five reasoning types (\textit{i.e.}, one-hop, conjunction, existence, multi-hop, negation). To reduce computational costs, we sample 1,000–1,500 instances from large test sets\footnote{Full-dataset experiments employing GPT-4o mini for both agents are provided in Appendix~\ref{sec:all_dataset}}.

\begin{table}[]
\centering
\renewcommand{\arraystretch}{1.2} 
\large 
\resizebox{\columnwidth}{!}{%
\begin{tabular}{c|c|c|c|c} 
\Xhline{0.7pt}
\textbf{Dataset}       & \textbf{Feature / Base} & \textbf{Answer Type}   & \begin{tabular}[c]{@{}c@{}}\textbf{Total} \# \\ \textbf{Test Set}\end{tabular} & \begin{tabular}[c]{@{}c@{}}\textbf{Used} \# \\ \textbf{Test Set}\end{tabular} \\ \hline
WebQSP        & Freebase      & Entity (M)    & 1639                                                               & 1639                                                              \\ 
CWQ       & Freebase      & Entity (M)    &   3531                                                             &   1000                                                            \\ 
MetaQA 3-hop  & Movie-related & Entity (M)    & 14274                                                              & 1000                                                              \\ 
FactKG        & DBpedia      & Boolean       & 9041                                                               & 1000                                                              \\ 
CRONQUESTIONS & Wikidata     & Entity/Number (S, M) & 16690                                                              & 1450                                                              \\ 
\Xhline{0.7pt}
\end{tabular}%
}
\caption{Dataset Statistics. (M): Multi-label QA, (S): Single-label QA.}
\label{tab:data-statistics}
\end{table}

\subsection{Baselines}

For comparison, we set KG-GPT \cite{kim-etal-2023-kg}, and ToG \cite{ToG} as baselines, as both can handle various KG structures and tasks to some extent. For fairness, all baselines are also evaluated with both low- and high-capacity LLMs, consistent with the R2-KG setup.
KG-GPT is a general framework adaptable for fact verification and QA tasks. However, it does not explicitly incorporate an \textit{abstention mechanism}, therefore we account for implicit \textit{Abstention} when it is unable to generate an answer due to token length constraints or formatting issues. Additionally, due to the structural modifications required to adapt KG-GPT for WebQSP and CWQ, we did not conduct experiments on this dataset.
ToG also employs an LLM agent for both KG exploration and answer generation. When ToG exceeds the depth limit (\textit{i.e.}, hop limit), it relies on the LLM’s parametric knowledge to generate answers, which we treat as \textit{Abstention}. However, we could not conduct an experiment for CRONQUESTIONS because ToG cannot handle time-structured KG queries without fundamental algorithmic changes. 
Additionally, we assess GPT-4o-mini’s ability to answer without KG access, treating predictions as correct if they convey the same meaning as the ground truth (\textit{e.g.}, \textit{`America'} equivalent with \textit{`USA'}).
For details on the modifications made to baselines, refer to Appendix ~\ref{sec:baseline_setting}.

\subsection{Experimental Setting} \label{subsec:experiment_setting}

For the \oag, we use six LLMs. We employ GPT-4o mini and GPT-4o \cite{openaigpt4omini, openaigpt4o} as API-based models, and LLaMA-3.1-70B-Instruct \cite{grattafiori2024llama3herdmodels}, Mistral-Small-Instruct-2409 \cite{mistral2023}, Qwen2.5-32B-Instruct, and Qwen2.5-14B-Instruct \cite{qwen2025qwen25technicalreport} as open-source LLMs.  
The maximum token length was set to $8,192$ for CRONQUESTIONS and FactKG, and $16,384$ for MetaQA, WebQSP, and CWQ.
Top-p and temperature were both set to $0.95$.
For the \sag, we use GPT-4o.
In the main experiment, \textit{T} was set to 15.
All experiments were conducted on a system equipped with two NVIDIA A100 GPUs and four NVIDIA RTX A6000 GPUs. Check models' spec in Appendix~\ref{sec:model_spec}.

\section{Main Results}
\begin{table*}[tb]
    \centering
	
        \resizebox{\textwidth}{!}{
    
		\begin{tabular}{@{}c|c|c|cccc|cccc|cccc|cccc|cc@{}}
			\toprule
			 \multirow{2}{*}{\rule{0pt}{12pt}\textbf{Method}} & \multicolumn{2}{c}{\textbf{Utilized Model}} & \multicolumn{4}{|c|}{\textbf{WebQSP}} & \multicolumn{4}{|c|}{\textbf{CWQ}} & \multicolumn{4}{|c|}{\textbf{MetaQA 3-hop}} & \multicolumn{4}{|c|}{\textbf{CRONQUESTIONS}} & \multicolumn{2}{c}{\textbf{FactKG}} \\ \cmidrule(l){2-3} \cmidrule(l){4-7} \cmidrule(l){8-11} \cmidrule(l){12-15} \cmidrule(l){16-19} \cmidrule(l){20-21}
		      & Operator & Supervisor & Cvg & F1 (M) & F1 (S) & Hit & Cvg & F1 (M) & F1 (S) & Hit & Cvg & F1 (M) & F1 (S) & Hit & Cvg & F1 (M) & F1 (S) & Hit & Cvg & Hit \\
            \midrule
            w/o KG & GPT-4o mini & --
                                 &  99.0 & 12.5 & 25.8 & 36.9 
                                 &  95.3 & 25.8 & 28.8 & 36.3 
                                 &  96.2 & 7.0  & 14.5 & 36.6 
                                 &  100  & 4.0  & 15.0 & 24.0 
                                 &  100  & 50.0 \\ 
            \midrule
            
            KG-GPT & Mistral-Small & --
                                 &  --   & --   & --   & --
                                 &  -- & -- & -- & -- 
                                 &  100  & 6.8  & 21.4 & 54.6 
                                 &  95.7 & 7.8  & 49.9 & 60.0
                                 &  55.4 & 57.6 \\
                                 
            KG-GPT & GPT-4o mini & --
                                 &  --   & --   & --   & --
                                 &  -- & -- & -- & -- 
                                 &  100  & 12.6 & 36.6 & 97.9 
                                 &  100  & 11.7 & 63.4 & 91.7
                                 &  100  & 63.3 \\
                                 
	      KG-GPT & GPT-4o & --      
                                 &  --   & --   & --   & --
                                 &  -- & -- & -- & -- 
                                 &  100  & 12.6 & 36.2 & 97.3
                                 &  100  & 10.6 & 60.3 & 83.8
                                 &  100  & 79.9 \\

            ToG & Mistral-Small & --    
                                 &  30.5 & 24.1 & 65.9 & 82.6
                                 &  15.4 & 48.2 & 57.8 & 64.3 
                                 &  24.1 & 13.2 & 31.2 & 62.2 
                                 &  --   & --   & --   & --
                                 &  52.8 & 69.5 \\
                                 
		  ToG & GPT-4o mini   & --    
                                 &  53.1 & 21.7 & 72.8 & \textbf{90.7}
                                 &  33.2 & 57.1 & 67.7 & 76.5 
                                 &  30.5 & 13.6 & 28.5 & 67.2 
                                 &  --   & --   & --   & --
                                 &  35.8 & 83.5 \\
                                 
            ToG & GPT-4o        & --
                                 &  58.8 & 21.9 & 69.6 & 89.1
                                 &  40.3 & 57.7 & 67.8 & 76.5 
                                 &  24.5 & 15.6 & 44.0 & 95.5 
                                 &  --   & --   & --   & --
                                 &  50.6 & 86.8 \\
            \midrule

            R2-KG & Qwen2.5-14B & GPT-4o    
                                                 &  76.4 & 75.7 & 80.9 & 87.9
                                                 &  51.5 & 73.6 & 76.7 & 82.3 
                                                 &  82.9 & 90.3 & 94.5 & 97.9
                                                 &  83.7 & \underline{40.4} & \underline{89.0} & 99.6
                                                 &  55.8 & \textbf{93.4} \\

            R2-KG & Qwen2.5-32B & GPT-4o    
                                                 &  81.5 & \textbf{79.4} & \textbf{83.0} & \underline{89.5}
                                                 &  59.4 & 69.3 & 77.7 & 82.8 
                                                 &  96.5 & \textbf{98.3} & \underline{99.1} & \textbf{100}
                                                 &  87.8 & 36.0 & 86.6 & \textbf{99.8}
                                                 &  64.1 & \underline{93.2} \\

            R2-KG & Mistral-Small & GPT-4o  
                                                 &  76.3 & 76.7 & \underline{82.3} & 89.4
                                                 &  40.3 & \textbf{76.9} & \textbf{79.8} & \textbf{85.1} 
                                                 &  75.0 & 94.5 & 96.3 & 99.3
                                                 &  65.9 & 33.1 & 87.6 & 99.4
                                                 &  43.2 & 93.1 \\
                                                 
            R2-KG & Llama-3.1-70B & GPT-4o         
                                                 &  81.0 & \underline{78.4} & 80.3 & 87.7
                                                 &  62.8 & \underline{75.6} & \underline{79.0} & \underline{84.2} 
                                                 &  94.9 & 97.7 & 98.7 & \underline{99.9}
                                                 &  84.1 & \textbf{42.2} & \textbf{89.9} & \underline{99.7}
                                                 &  57.3 & 92.7 \\

            R2-KG & GPT-4o mini & GPT-4o    
                                                 &  81.3 & 73.6 & 80.1 & 88.4
                                                 &  63.1 & 69.6 & 77.6 & 82.4 
                                                 &  94.6 & 95.7 & 97.6 & \underline{99.9}
                                                 &  90.4 & 34.3 & 85.6 & 99.4
                                                 &  70.2 & 92.5 \\

            R2-KG & GPT-4o & GPT-4o
                                                 &  85.3 & 71.1 & 81.4 & 89.1
                                                 &  76.2 & 71.2 & 76.7 & 82.3 
                                                 &  98.3 & \textbf{98.3} & \textbf{99.2} & \underline{99.9}
                                                 &  90.8 & 33.6 & 85.3 & 99.5
                                                 &  77.8 & 93.1 \\
                                                 
            \bottomrule
        \end{tabular}%
        }
        
    \caption{Performance of baselines and \modelname on the five KG-based reasoning benchmarks. We denote the \textbf{best} and \underline{second-best} method for each metric (except coverage). 
    Cvg: Coverage, F1 (M): Micro F1 score, F1 (S): Samplewise F1 score.
    }
    \label{tab:2agent_Operator_result}
\end{table*}%

\subsection{Performance of \modelname}
As shown in Table~\ref{tab:2agent_Operator_result}, \modelname consistently outperforms baselines in F1 score and achieves higher hit rates on four out of five benchmarks. Even with low-capacity LLMs as the \oag, \modelname surpasses ToG and KG-GPT, which fully rely on GPT-4o (\textit{i.e.}, high-capacity LLM) throughout the reasoning process. Additionally, \modelname achieves a hit rate of over 90\% in three out of the five benchmarks, with MetaQA 3-hop reaching 100\%. On WebQSP, ToG with GPT-4o mini marginally outperforms \modelname in terms of hit rate, but \modelname achieves significantly higher F1 scores, which is a more suitable metric for multi-label QA, demonstrating its superior reasoning performance. 
This highlights the advantage of \modelname not only in single-label QA but also in multi-label QA.
WebQSP and CWQ yielded hit rates below 90\%, CRONQUESTIONS showed micro F1 under 43\% across all models—a pattern further analyzed in Appendix~\ref{sec:upperbound_reason}.
The strong performance of \modelname can be attributed to its \oag{}’s ability to accumulate and utilize information from previous hops in multi-hop reasoning. Within a given \textit{T}, the framework can revisit and adjust incorrect paths from prior steps, dynamically selecting alternative paths as needed. Furthermore, during inference, the \sag is not restricted to a single reasoning path but can flexibly combine relevant triples, leading to more accurate reasoning and answer generation.

To ensure that \modelname’s performance does not rely on the capability of high-capacity LLMs, we conducted additional experiments by varying not only the \oag but also the \sag across different model scales. The results show that even when the Supervisor is a low-capacity LLM, \modelname still achieves higher F1 scores and hit rates than the baseline. Please check the Appendix~\ref{sec:op_sup_same_model}.

\subsection{Coverage Across Different LLMs}
Note that \modelname's coverage is the highest across all cases when using GPT-4o as the \oag.
When using relatively low-capacity LLMs, the coverage decreases in varying degrees.
The reason why high-capacity LLMs as a \oag achieve higher coverage is twofold: First, they excel at collecting key evidence, allowing them to request \textit{Verification(\(\cdot\))} at the optimal moment. 
Second, their strong language understanding enables them to effectively use the feedback provided by the \sag. Table~\ref{tab:2agent_Operator_result} shows that even with a low-capacity LLM as the \oag, \modelname maintains a high F1 score and hit rate despite reduced coverage. This highlights the advantage of \modelname's separation of the \oag and \sag. Since \modelname maintains answer reliability while only affecting coverage, users can confidently choose an \oag based on their budget constraints.

\subsection{Case analysis of Abstention}
Even when \textit{T} is high, reasoning may still fail, leading to abstention.
The most common cases are: (1) Repeated helper function requests—The \oag redundantly calls the same function across multiple steps, even after retrieving the necessary information in previous steps. 
(2) Failure to interpret \sag{}'s feedback—The \oag struggles to incorporate the \sag{}'s instructions, especially when directed to collect additional information about a specific entity’s relation, failing to refine exploration in later steps.
(3) Failure to extract an answer despite sufficient evidence—When the retrieved triple set is overly large, the \sag may misinterpret relationships between triples, leading to incorrect judgment.
(4) Incorrect function call format—The \oag does not follow the predefined format when calling a helper function, causing parsing issues that prevent information retrieval.

\subsection{LLM Usage Comparison}
Table~\ref{tab:call_statistics} shows that \oag and \sag average 5.94–8.63 and 1.04–1.43 calls per \textit{q}, respectively; by contrast, KG-GPT makes at least 3 high-capacity LLM calls, and ToG makes minimum 4 to maximum 25 such calls in the whole reasoning process. \modelname employs \textit{Low/High-Capacity LLM Separation} for accuracy and cost efficiency, significantly reducing high-capacity LLM usage to an average of 1.28 calls per \textit{q}, making it both cost-effective and superior in performance. 


\begin{table}[ht]
\scriptsize 
\centering
\resizebox{\columnwidth}{!}{%
\begin{tabular}{c|c|c}
\Xhline{0.7pt} 
\textbf{Dataset}       & \textbf{Operator Call} & \textbf{Supervisor Call} \\ \hline
WebQSP        & 5.94          & 1.04 \\
CWQ        &  8.38         & 1.28 \\
MetaQA 3-hop        & 8.63          & 1.38 \\
FactKG        & 8.21          & 1.43 \\
CRONQUESTIONS & 7.34          & 1.27 \\
\Xhline{0.7pt} 
\end{tabular}%
}
\caption{Number of LLM calls per sample for \textit{Operator} and \textit{Supervisor} in different datasets}
\label{tab:call_statistics}
\end{table}

\begin{table}[t]
\centering
\small
\begin{tabular}{lccc}
\hline
\textbf{Method} & \textbf{Operator} & \textbf{Supervisor} & \textbf{Total} \\
\hline
Dual-agent & 9.89\,s & 3.17\,s & 13.06\,s \\
Single-agent & 12.47\,s & -- & 12.47\,s \\
\hline
\end{tabular}
\caption{Time spent per method on CWQ}
\label{tab:cwq_time}
\end{table}

\begin{table}[t]
\centering
\small
\begin{tabular}{lcc}
\hline
\textbf{Method} & \textbf{Operator} & \textbf{Supervisor} \\
\hline
Dual-agent & 18{,}987 & 4{,}912 \\
Single-agent & 19{,}309 & -- \\
\hline
\end{tabular}
\caption{Token usage per method on CWQ}
\label{tab:cwq_token}
\end{table}

\section{Further Analysis}

\begin{figure*}[ht]
    \centering
    \includegraphics[width=0.95\textwidth, height=4.2cm]{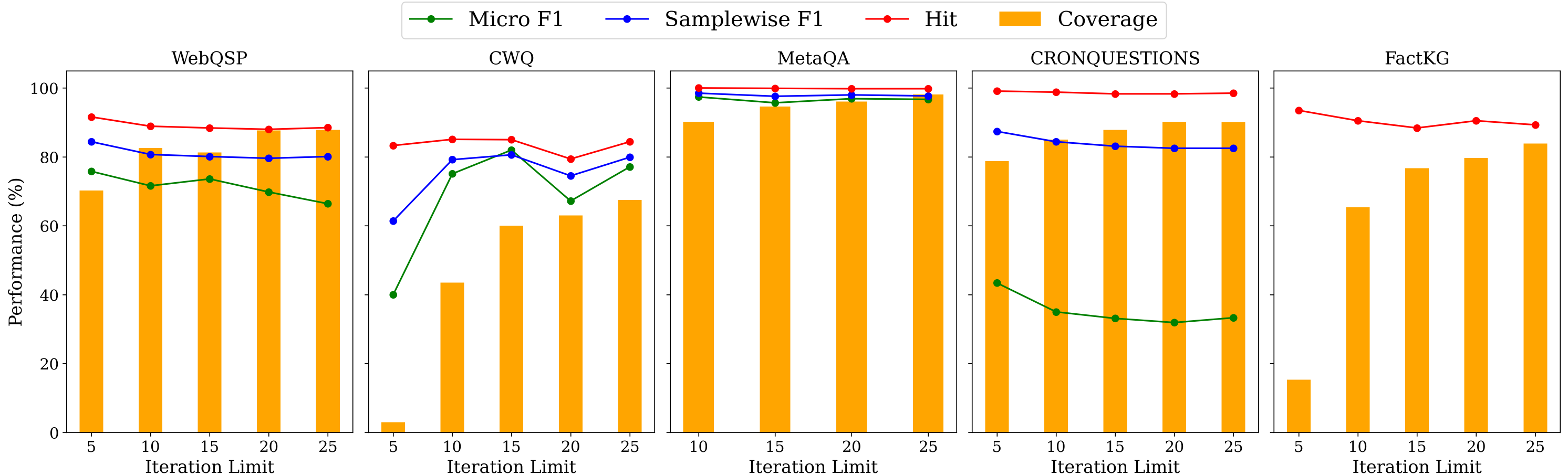}
    \caption{Changes in coverage, F1 Scores, and hit rate based on Iteration Limit}
    \label{fig:iter-limit}
\end{figure*}

\subsection{Effect of Iteration Limit}
Figure~\ref{fig:iter-limit} illustrates the impact of \textit{T} on coverage, F1 scores, and hit rate. At $5\leq T\leq15$, coverage improves, whereas F1 scores and hit rate slightly decline. Lower \(T (=5)\) causes early termination, leading to lower coverage but higher accuracy on simpler queries (exception arises in CWQ—due to extremely low coverage, few errors affect overall performance).
At $10\leq T\leq15$, increased evidence collection enhances coverage, though accuracy slightly drops as queries grow more complex.
Beyond 20 iterations, coverage stabilizes while F1 scores and hit rates marginally decrease. This suggests that the optimal iteration range is 10-15 for the benchmarks we used, as further steps mainly introduce redundant exploration that is unhelpful for reasoning.

\begin{table*}[tb]
    \centering
        \resizebox{\textwidth}{!}{
    
            \begin{tabular}{@{}c|c|cccc|cccc|cccc|cccc|cc@{}}
			\toprule
			 \multirow{2}{*}{\rule{0pt}{18pt}\textbf{\makecell{Reasoning Path \\ Strategy}}} & \multicolumn{1}{c}{\textbf{Utilized Model}} & \multicolumn{4}{|c|}{\textbf{WebQSP}} & \multicolumn{4}{|c|}{\textbf{CWQ}} & \multicolumn{4}{|c|}{\textbf{MetaQA 3-hop}} & \multicolumn{4}{|c|}{\textbf{CRONQUESTIONS}} & \multicolumn{2}{c}{\textbf{FactKG}} \\ \cmidrule(l){2-2} \cmidrule(l){3-6} \cmidrule(l){7-10} \cmidrule(l){11-14} \cmidrule(l){15-18} \cmidrule(l){19-20} 
		      &  Operator & Cvg & F1 (M) & F1 (S) & Hit & Cvg & F1 (M) & F1 (S) & Hit & Cvg & F1 (M) & F1 (S) & Hit & Cvg & F1 (M) & F1 (S) & Hit & Cvg & Hit \\
            \midrule
                                \multirow{5}{*}{Multi Prompts}
                                     & Qwen2.5-14B  
                                     &  54.0 & 70.4 & 80.4 & 92.4 
                                     &  34.7 & 73.3 & 77.8 & 85.9 
                                     &  -- & -- & -- & -- 
                                     &  69.7 & 26.7 & 79.0 & 97.4 
                                     &  44.5 & 94.4 \\ 

                                     & Qwen2.5-32B 
                                     &  62.1 & 69.2 & 81.2 & 93.6 
                                     &  40.9 & 71.9 & 80.0 & 87.5 
                                     &  80.8 & \underline{95.7} & \underline{96.3} & \textbf{100}  
                                     &  78.2 & 25.8 & 75.9 & \textbf{98.9} 
                                     &  67.3 & 92.3 \\ 

                                     & Mistral-Small 
                                     &  40.5 & \underline{76.3} & \underline{84.0} & \textbf{95.9} 
                                     &  21.4 & 66.9 & 75.7 & 85.0 
                                     &  58.1 & 86.4 & 88.5 & \textbf{100}  
                                     &  39.0 & \underline{47.7} & \textbf{86.8} & 98.6 
                                     &  25.5 & 93.3 \\ 

                                     & Llama-3.1-70B 
                                     &  57.2 & 74.1 & 81.6 & 93.3 
                                     &  46.5 & \textbf{79.9} & \underline{82.5} & 87.5 
                                     &  86.7 & 93.4 & 94.7 & \textbf{100} 
                                     &  73.9 & 30.0 & \underline{84.6} & \textbf{98.9} 
                                     &  58.2 & 93.8 \\ 
                                     
                                     & GPT-4o mini 
                                     &  65.1 & 71.9 & 83.3 & 93.8 
                                     &  36.0 & 74.3 & 80.0 & 87.5 
                                     &  84.1 & 90.9 & 95.2 & \textbf{100} 
                                     &  64.3 & 18.0 & 72.5 & 96.8 
                                     &  76.1 & 92.1 \\ 

                                \cline{1-20}\rule{0pt}{10pt}
                                \multirow{5}{*}{Paraphrasing} 
                                     & Qwen2.5-14B 
                                     &  54.7 & 61.0 & 78.0 & 94.0 
                                     &  33.4 & 71.8 & 76.1 & 83.8 
                                     &  -- & -- & -- & -- 
                                     & 54.3  & 40.6 & 81.7 & 97.1 
                                     &  48.6 & 94.2\rule{0pt}{12pt} \\ 

                                     & Qwen2.5-32B 
                                     &  68.4 & 64.8 & 78.4 & 92.6 
                                     &  39.9 & 75.3 & 78.4 & 85.5 
                                     &  77.7 & 94.8 & 95.7 & 99.9 
                                     &  95.6 & 27.3 & 76.7 & 97.5 
                                     &  47.9 & \textbf{95.3} \\ 

                                     & Mistral-Small 
                                     &  50.5 & 70.3 & 81.4 & 93.8 
                                     &  19.0 & 63.5 & 72.1 & 82.6 
                                     &  60.4 & 82.5 & 84.8 & 99.7 
                                     & 58.1  & 47.5 & 84.3 & 98.2 
                                     &  24.6  & 95.1 \\ 

                                     & Llama-3.1-70B 
                                     &  57.0 & 75.3 & 82.8 & 94.4 
                                     &  44.4 & 79.4 & 81.9 & 87.8 
                                     &  89.5 & \textbf{95.8} & 96.1 & \textbf{100} 
                                     & 78.8  & 37.5  & 73.1 & 97.7 
                                     &  50.2 & 94.8 \\ 
                                     
                                     & GPT-4o mini 
                                     &  69.6 & 60.5 & 81.2 & 92.5 
                                     &  44.0 & 69.1 & 80.3 & \textbf{88.6} 
                                     &  82.4 & 92.9  & 95.8 & \textbf{100} 
                                     & 94.1  &  29.4 & 80.1 & 96.7 
                                     &  71.7 & 92.5 \\ 

                                \cline{1-20}\rule{0pt}{10pt}
                                \multirow{5}{*}{\makecell{Top-p /\\Temperature}}

                                     & Qwen2.5-14B 
                                     &  51.3 & 67.6 & 80.1 & 94.3 
                                     &  36.4 & 73.8 & 78.8 & 86.5 
                                     &  -- & -- & -- & -- 
                                     & 60.0 & 46.1  & 84.0 & 96.9 
                                     &  42.2 & 92.9\rule{0pt}{12pt} \\ 

                                     & Qwen2.5-32B 
                                     &  67.5 & 63.8 & 78.9 & 91.9 
                                     &  41.2 & 71.3 & 78.5 & 85.9 
                                     &  74.5 & 93.0 & 94.9 & 99.9 
                                     & 13.3 & 12.7  & 29.7 & 92.7 
                                     &  48.9 & 92.9 \\ 

                                     & Mistral-Small 
                                     &  51.4 & 75.3 & \textbf{84.2} & \underline{95.0} 
                                     &  14.1 & 59.9 & 71.4 & 83.7 
                                     &  61.0 & 87.6 & 89.8 & 99.8 
                                     & 71.9  & 38.1 & 84.1 & 98.5 
                                     &  15.7 & 90.4 \\ 

                                     & Llama-3.1-70B 
                                     &  63.1 & \textbf{79.9} & 82.5 & 93.1 
                                     &  49.1 & 78.6 & 81.0 & 86.4 
                                     &  91.0 & 95.2 & \textbf{96.5} & \textbf{100} 
                                     & 94.4  & \textbf{49.2} & 83.2 & 97.2 
                                     &  58.6 & 93.3 \\ 
                                     
                                     & GPT-4o mini 
                                     &  69.1 & 61.9 & 79.8 & 91.1 
                                     &  42.5 & \underline{79.8} & \textbf{83.0} & \underline{88.2} 
                                     &  84.4 & 90.4 & 94.1 & 99.2 
                                     & 95.9 & 31.2 & 80.1 & 96.8 
                                     &  37.9 & \textbf{95.3} \\ 

            \bottomrule
        \end{tabular}%
        }

    \caption{Performance of single-agent version of \modelname with self-consistency on the five KG-based reasoning benchmarks. We denote the \textbf{best} and \underline{second-best} method for each metric (except coverage). 
    Cvg: Coverage, F1 (M): Micro F1 score, F1 (S): Samplewise F1 score.
    }
    \label{tab:ablation_ensemble_result}

\end{table*}%
\subsection{Single-Agent Version of \modelname with Strict Self-Consistency} \label{sec:ensemble_approach}
To further reduce the cost of using a high-capacity LLM as the \sag, we leverage a self-consistency \cite{wang2023selfconsistencyimproveschainthought} strategy where the \oag handles both evidence collection and answer generation (\textit{i.e.}, single-agent version of \modelname). Without the \sag, the \oag assesses evidence sufficiency and generates answers within \textit{Verification(\(\cdot\))}. The reasoning process runs three trials per instance with $T=10$, following these rules;  
First, unlike the typical majority-based self-consistency strategy, our approach enforces a stricter unanimous agreement criterion for the final prediction. 
Second, if no agreement is reached or if \textit{Abstention} appears in any attempt, the final prediction is also \textit{Abstention}.  
We apply three reasoning path strategies;  
Multi-Prompts (distinct in-context examples for the same query),  
Query Paraphrasing (semantically equivalent query variations),  
Top-p/Temperature Variation (sampling diversity)\footnote{Detailed experimental settings and prompt examples are provided in Appendix~\ref{sec:single_agent_method}}.

Table~\ref{tab:ablation_ensemble_result} shows a significant decrease in coverage compared to the dual-agent version of \modelname, while F1 scores and hit rate were comparable or slightly improved except for MetaQA 3-hop and CRONQUESTIONS. Despite this, it still significantly outperformed baselines, achieving 100\% on MetaQA 3-hop and micro F1 gains (WebQSP +55.8\%, CWQ +22.2\%, MetaQA 3-hop +80.2\%, and CRONQUESTIONS +37.5\%) compared to the baselines.
These results demonstrate that a single-agent variant of \modelname can achieve higher-than-baseline answer reliability at even lower cost. Multi-Prompts generally showed strong performance across all datasets.
However, relying solely on low-capacity LLMs limits adaptability to more complex KGs like CRONQUESTIONS (\textit{i.e.,} KGs that require reasoning over temporal constraints and time-sensitive relations), and stricter filtering inevitably leads to reduced coverage and overall utility compared to the dual-agent setup.

\subsection{Comparison of Time Consuming and Token Usage of Dual/Single Version of R2-KG}\label{sec:token_usage}
In the dual-agent setup, the \oag calls the \textit{Verification(\(\cdot\))} function to request the \sag’s help in gathering sufficient evidence. In the single-agent variant, however, the \oag must generate an answer directly by referencing the provided triples when \textit{Verification(\(\cdot\))} is invoked. We measured the average time spent and token usage on CWQ with GPT-4o mini as the \oag and GPT-4o as the \sag. Table~\ref{tab:cwq_time} shows that the latency gap between the dual-agent and single-agent versions is under one second. This represents only a minor portion of the total query-processing time, supporting our claim that the dual-agent framework’s latency is not practically problematic. 
We also compared token usage between the dual-agent and single-agent settings. As shown in Table~\ref{tab:cwq_token}, the dual-agent framework used a total of 18,987 tokens for the \oag and 4,912 for the \sag, while the single-agent version used 19,309 tokens in total—resulting in a difference of 4,590 tokens. Restricting the high-capacity LLM to the supervisor role in the dual-agent version markedly lowers token usage compared to the single-agent version. The advantage of the dual-agent framework is therefore evident, as it guarantees both high coverage and reliability. Nonetheless, by using low-capacity LLM as a \oag, the single-agent version with self-consistency remains a practical alternative for cost-sensitive applications; although its coverage is somewhat reduced, it still ensures high reliability, making it a viable option when cost is the primary constraint.

\section{Conclusion}

We propose \modelname, the first general KG-based reasoning framework with an \textit{abstention mechanism}, ensuring the reliability for various KG-based reasoning tasks.  
Separation of \oag and \sag reduced high-capacity LLM usage, leading to a cost-effective solution for KG-based reasoning. 
Moreover, in simpler KGs, the single-agent version of \modelname with strict self-consistency can maintain reliability while further reducing cost.

\section*{Limitations}
The \sag makes the final prediction based solely on the triple set and relation list collected by the \oag. Consequently, it cannot determine whether the retrieved information is minimal or exhaustive. In multi-label QA tasks, this limitation may cause underprediction, where the framework generates fewer answers than the actual number of correct labels.  
Additionally, if a query can be answered through multiple relation paths, the \sag may provide an answer as long as one valid path exists, potentially overlooking alternative correct paths. One way to mitigate this would be to involve the \sag in every iteration step, but this would remove the distinction between the \oag and \sag roles, increasing computational costs.  
These constraints stem from the trade-off between cost-effectiveness and reasoning efficiency. While the current design optimizes resource usage, it may not always capture all possible answers in complex reasoning scenarios.

\section*{Ethical Consideration}
LLM-based KG reasoning requires substantial computational resources, which can contribute to environmental concerns. While our study proposes methods to reduce overall LLM usage, the reliance on large-scale models remains a consideration in terms of environmental impact.

\section*{Acknowledgements}
This work was supported by the Institute of Information \& Communications Technology Planning \& Evaluation (IITP) grants (No.RS-2019-II190075, No.RS-2022-II220984, No.RS-2024-00338140) and National Research Foundation of Korea (NRF) grants (NRF-2020H1D3A2A03100945), funded by the Korea government (MSIT).

\bibliography{acl_latex}

\appendix
\newpage

\begin{table*}[]
\resizebox{\textwidth}{!}{%
\begin{tabular}{c|c|c} 
\hline
Model                                   & Parameter                   & Architecture                                                                                         \\ \hline
Qwen2.5-14B                              & 14.7B                        & transformers with RoPE, SwiGLU, RMSNorm, and Attention QKV bias                 \\
Qwen2.5-32B                              & 32B                          & transformers with RoPE, SwiGLU, RMSNorm, and Attention QKV bias                 \\
Llama-3.1-70B                            & 70B                          & auto-regressive language model that uses an optimized transformer architecture. \\
Mistral-Small-Instruct-2409              & 22B                          & Unknown                                                                         \\
GPT-4o mini, GPT-4o & \multicolumn{1}{l|}{Unknown} & Unknown                                                                         \\ \hline
\end{tabular}%
}
\caption{Specification of models}
\label{sec:llm_spec}
\end{table*}

\begin{table}[]
    \small
    \centering
    \resizebox{\columnwidth}{!}{
            \begin{tabular}{@{}c|cccc@{}}
			\toprule
			 \multirow{2}{*}{\rule{0pt}{13pt}\textbf{Dataset}} & \multicolumn{4}{c}{\textbf{Metrics}} \\  \cmidrule(l){2-5}
		      & Coverage & F1 (M) & F1 (S) & Hit \\
            \midrule
                                     MetaQA 1-hop
                                     &  100 & 95.3 & 98.9 & 100 \\  

                                \midrule
                                     MetaQA 2-hop
                                     &  96.0 & 99.9 & 99.7 & 100 \\ 

            \bottomrule
        \end{tabular}%
        }

    \caption{Performance of \modelname in MetaQA 1-hop and 2-hop.
     F1 (M): Micro F1 score, F1 (S): Samplewise F1 score.
    }
    \label{tab:appendix_metaqa}

\end{table}%

\begin{table*}[tb]
    \centering
        \resizebox{\textwidth}{!}{
    
            \begin{tabular}{@{}c|c|c|cccc|cccc|cccc|cc@{}}
			\toprule
			 \multirow{2}{*}{\rule{0pt}{18pt}\textbf{\makecell{Reasoning Path \\ Strategy}}} & \multicolumn{2}{c}{\textbf{Utilized Model}} & \multicolumn{4}{|c|}{\textbf{WebQSP}} & \multicolumn{4}{|c|}{\textbf{MetaQA 3-hop}} & \multicolumn{4}{|c|}{\textbf{CRONQUESTIONS}} & \multicolumn{2}{c}{\textbf{FactKG}} \\ \cmidrule(l){2-3} \cmidrule(l){4-7} \cmidrule(l){8-11} \cmidrule(l){12-15} \cmidrule(l){16-17} 
		      &  Operator & Supervisor & Coverage & F1 (M) & F1 (S) & Hit & Coverage & F1 (M) & F1 (S) & Hit & Coverage & F1 (M) & F1 (S) & Hit & Coverage & Hit \\
            \midrule
                                Multi Prompts 
                                     & GPT-4o mini & GPT-4o
                                     &  69.2 & 77.1 & 85.5 & 93.6 
                                     &  89.2 & 92.3 & 95.4 & \textbf{100}  
                                     & 83.6  & \textbf{34.5}& \textbf{85.1} & 98.6 
                                     &  56.3 & 94.5 \\ 

                                \midrule
                                Paraphrasing 
                                     & GPT-4o mini & GPT-4o
                                     &  70.7 & 73.8 & 85.7 & \textbf{93.9} 
                                     &  87.3 & \textbf{92.5} & \textbf{95.9} & 99.9 
                                     & 84.0  & 33.5 & 84.2 & 98.4 
                                     &  55.0 & 93.6 \\ 

                                \midrule
                                \makecell{Top-p /\\Temperature} 
                                     & GPT-4o mini & GPT-4o
                                     &  73.6 & \textbf{81.2} & \textbf{86.0} & 92.8 
                                     &  87.1 & 92.4 & 95.8 & \textbf{100}  
                                     & 23.9  & 27.8 & 77.4 & \textbf{99.7} 
                                     &  16.6 & \textbf{95.2} \\ 

            \bottomrule
        \end{tabular}%
        }

    \caption{Performance of dual-agent version of \modelname with self-consistency on the four KG-based reasoning benchmarks. We denote the \textbf{best} method for each metric (except coverage). 
     F1 (M): Micro F1 score, F1 (S): Samplewise F1 score.
    }
    \label{tab:ablation_Supervisoragent_ensemble_result}

\end{table*}%
\begin{table*}[tb]
    \centering
	
        \resizebox{\textwidth}{!}{
    
		\begin{tabular}{@{}c|c|cccccccccccccc@{}}
			\toprule
			 \multicolumn{2}{c}{\textbf{Utilized Model}} & \multicolumn{4}{|c}{\textbf{WebQSP}} & \multicolumn{4}{c}{\textbf{MetaQA 3-hop}} & \multicolumn{4}{c}{\textbf{CRONQUESTIONS}} & \multicolumn{2}{c}{\textbf{FactKG}} \\ \cmidrule(l){1-2} \cmidrule(l){3-6} \cmidrule(l){7-10} \cmidrule(l){11-14} \cmidrule(l){15-16} 
		      Operator & Supervisor & Coverage & F1 (M) & F1 (S) & Hit & Coverage & F1 (M) & F1 (S) & Hit & Coverage & F1 (M) & F1 (S) & Hit & Coverage & Hit \\
            \midrule
               
            GPT-4o mini & GPT-4o mini 
                                                 &  85.0 & 78.8 & 79.9 & 88.3
                                                 &  65.7 & 92.8 & 95.7 & 98.6
                                                 &  84.9 & 30.5 & 81.8 & 98.8
                                                 &  78.4 & 93.0 \\
                                                 
            \bottomrule
        \end{tabular}%
        }
        
    \caption{Performance of baselines and \modelname on the four KG-based reasoning benchmarks on the entire test set.
    F1 (M): Micro F1 score, F1 (S): Samplewise F1 score.
    }
    \label{tab:supplement_allmini}
\end{table*}%

\begin{table}[]
    \small
    \centering
        \resizebox{\columnwidth}{!}{
    
            \begin{tabular}{@{}c|c|c|cccc@{}}
			\toprule
			 \multirow{2}{*}{\rule{0pt}{14pt}\textbf{\makecell{Method}}} & \multicolumn{2}{c}{\textbf{Utilized Model}} & \multicolumn{4}{|c}{\textbf{Synthetic MetaQA 3-hop}} \\ \cmidrule(l){2-3} \cmidrule(l){4-7}
		      &  Operator & Supervisor & Cvg & F1 (M) & F1 (S) & Hit \\
            \midrule
                                \multirow{3}{*}{KG-GPT} 
                                     & Mistral-Small & --
                                     &  100 & 4.5 & 16.1 & 42.0 
                                     \\
                                      & GPT-4o mini & --
                                     &  99.0 & 10.8 & 27.0 & 94.9 
                                     \\
                                      & GPT-4o & --
                                     &  99.0 & 10.9 & 27.8 & 96.0 
                                     \\

                                \midrule
                                \multirow{3}{*}{ToG} 
                                     & Mistral-Small & --
                                     &  19.5 & 3.3 & 7.8 & 28.2 
                                     \\
                                      & GPT-4o mini & --
                                     &  15.0 & 14.2 & 19.5 & 70.0 
                                     \\
                                      & GPT-4o & --
                                     &  14.5 & 15.3 & 21.1 & 89.7 
                                     \\

                                \midrule
                                \multirow{6}{*}{R2-KG} 
                                     & Qwen2.5-14B & GPT-4o
                                     &  88.0 & 89.1 & 89.6 & 93.8 
                                     \\
                                     & Qwen2.5-32B & GPT-4o
                                     &  90.5 & \textbf{98.4} & 98.8 & 99.4 
                                     \\
                                     & Mistral-Small & GPT-4o
                                     &  58.5 & 92.4 & 93.4 & 94.9 
                                     \\
                                     & Llama-3.1-70B & GPT-4o
                                     &  87.0 & 94.8 & 96.9 & 98.9 
                                     \\
                                      & GPT-4o mini & GPT-4o
                                     &  87.5 & 93.0 & 94.1 & 96.6 
                                     \\
                                      & GPT-4o & GPT-4o
                                     &  97.0 & 98.1 & \textbf{99.1} & \textbf{100} 
                                     \\

            \bottomrule
        \end{tabular}%
        }

    \caption{Performance using synthetic KG (modified KG based on MetaQA). We denote the \textbf{best} method for each metric (except coverage).  
    Cvg: Coverage, F1 (M): Micro F1 score, F1 (S): Samplewise F1 score.
    }
    \label{tab:metaqa_synthetic}

\end{table}%

\section{Performance on 1-Hop and 2-Hop Questions} \label{sec:metaqa_12hop}
The MetaQA dataset consists of 1-hop, 2-hop, and 3-hop questions; however, our experiments focused exclusively on 3-hop questions. Given that the KG of MetaQA is relatively small and that 1-hop and 2-hop questions are considerably simpler than 3-hop questions, we excluded them from our primary evaluation. Nevertheless, to assess the \modelname across different levels of task complexity, we randomly sampled 100 questions from 1-hop and 2-hop sets and evaluated the performance. As shown in Table~\ref{tab:appendix_metaqa}, \modelname exhibited strong performance with high coverage.

\section{Examples of the Two Excluded Question Types in CRONQUESTIONS}\label{sec:cronq_excluded}
Unlike the four other datasets, CRONQUESTIONS is constructed with a five-element KG, where each quintuple follows the format:
[head, relation, tail, start time, end time]. This structure includes temporal information, specifying the start and end years of an event.
CRONQUESTIONS contains five types of reasoning tasks: Simple time, Simple entity, Before/After, First/Last, and Time Join.
However, in our experiments, we excluded the Before/After and First/Last question types. The primary reason is that, while our framework predicts answers based on the KG, these question types often contain subjective ground truth labels that do not fully align with the available KG information. For example, this is a sample of Before/After question:
``Which team did Roberto Baggio play for before the Italy national football team?''
Using our framework, we can retrieve the following KG facts related to Roberto Baggio:
[Roberto Baggio, member of sports team, ACF Fiorentina, 1985, 1990]
[Roberto Baggio, member of sports team, Brescia Calcio, 2000, 2004]
[Roberto Baggio, member of sports team, Vicenza Calcio, 1982, 1985]
[Roberto Baggio, member of sports team, Juventus F.C., 1990, 1995]
[Roberto Baggio, member of sports team, Italy national football team, 1988, 2004]
[Roberto Baggio, member of sports team, A.C. Milan, 1995, 1997]
[Roberto Baggio, member of sports team, Bologna F.C. 1909, 1997, 1998].
According to the KG, Roberto Baggio joined the Italy national football team in 1988. Before that, he played for Vicenza Calcio (starting in 1982) and ACF Fiorentina (starting in 1985), meaning both teams are valid answers. 
However, the ground truth label in the dataset only includes ACF Fiorentina, omitting Vicenza Calcio, despite it being a correct answer based on the KG. Due to this labeling inconsistency, objective evaluation of these question types becomes unreliable. As a result, we decided to exclude these types from our experiments.

\begin{table}[t]
\centering

\setlength{\arrayrulewidth}{0.2pt}

\begingroup
\setlength{\arrayrulewidth}{0.1pt} 
\scriptsize                      
\fontseries{l}\selectfont
\renewcommand{\arraystretch}{0.8} 
\resizebox{\columnwidth}{!}{
\begin{tabular}{llcc}
\toprule
\multicolumn{4}{c}{\textbf{FactKG}} \\
\midrule
\textbf{Method} & \textbf{Model} & \textbf{Cvg} & \textbf{Hit} \\
\midrule
ToG & Mistral-Small & 52.8 & 69.5 \\
ToG & GPT-4o mini & 35.8 & 83.5 \\
ToG & GPT-4o & 50.6 & 86.8 \\
\midrule
R2-KG & Mistral-Small & 56.2 & 82.9 \\
R2-KG & GPT-4o mini & 78.4 & 89.1 \\
R2-KG & GPT-4o & 77.8 & 93.1 \\

\bottomrule
\end{tabular}
}
\endgroup

\normalsize
\renewcommand{\arraystretch}{1.0}

\vspace{0.5em}

\normalsize
\setlength{\arrayrulewidth}{0.2pt}
\resizebox{\columnwidth}{!}{%
\begin{tabular}{llcccc}
\toprule
\multicolumn{6}{c}{\textbf{MetaQA 3-hop}} \\
\midrule
\textbf{Method} & \textbf{Model} & \textbf{Cvg} & \textbf{F1(M)} & \textbf{F1(S)} & \textbf{Hit} \\
\midrule
ToG & Mistral-Small & 24.1 & 13.2 & 31.2 & 62.2 \\
ToG & GPT-4o mini & 30.5 & 13.6 & 28.5 & 67.2 \\
ToG & GPT-4o & 24.5 & 15.6 & 44.0 & 95.5 \\
\midrule
R2-KG & Mistral-Small & 79.0 & 77.9 & 84.3 & 92.7 \\
R2-KG & GPT-4o mini & 65.7 & 92.8 & 95.7 & 98.6 \\
R2-KG & GPT-4o & 98.3 & 98.3 & 99.2 & 99.9 \\

\bottomrule
\end{tabular}
}

\vspace{0.4em}
\caption{Results on FactKG and MetaQA 3-hop benchmarks.
R2-KG employs both \oag and \sag agents with three different models.}
\label{2agent_main_ver2}
\end{table}

\section{\modelname vs. baselines with various LLMs as \oag and \sag}\label{sec:op_sup_same_model}
To demonstrate that \modelname consistently outperforms the baseline regardless of the underlying LLM’s capability, we conducted the following experiment. In the dual-agent setup, we varied not only the \oag but also the \sag across three different LLMs(LLMs—Mistral-Small, GPT-4o mini, and GPT-4o). The results show that even when the \sag is a low-capacity LLM, \modelname still achieves higher F1 scores and hit rates than the baseline. As shown in Table~\ref{2agent_main_ver2}, on both the FactKG and MetaQA 3-hop datasets, \modelname consistently outperforms ToG by a substantial margin. In FactKG, when both the \oag and \sag use GPT-4o mini, \modelname achieves a hit rate of 89.1\%, whereas ToG with GPT-4o reaches only 86.8\%. This clearly demonstrates that even when using a low-capacity model, \modelname surpasses the baseline equipped with a high-capacity model. Similarly, on MetaQA 3-hop, \modelname outperforms ToG by nearly 30\% in terms of hit rate, further confirming its robustness and efficiency across model scales.

\section{Prompt Structure and Usage}\label{sec:prompt}
Each prompt for \oag consists of three components:
Task description, Helper function explanations, and three few-shot examples.
When using \modelname, users only need to modify the few-shot examples to match the specific dataset while keeping the rest of the prompt unchanged. Examining the prompts reveals that when the \oag requests a helper function, the \oag can request multiple instances in a single iteration based on its needs. Additionally, it can request different types of functions simultaneously. The prompt for \sag contains the following elements: Task description, triples collected so far by the \oag, a relation list for each entity, and few-shot examples.
The reason for explicitly including the entity-wise relation list is to ensure that when the \sag provides feedback to the \oag, it requests subgraphs that actually exist in the KG. During pilot testing, when the relation list was not provided, the system occasionally requested non-existent entity-relation pairs in the KG. This resulted in ineffective feedback and ultimately failed to assist the \oag in its KG exploration.

\section{Final Reasoning Path Construction}\label{sec:finalgoldpath}
When sufficient evidences are given from \oag to \sag, then the \sag selects the necessary triples and constructs a final reasoning path that aligns with the claim structure.
Assume that the given query is ``Which languages were used in the films directed by the same directors as [The Vanishing American]'' and \(G_k\) given by the \oag are as follows (tilde (\textasciitilde)\space represents the inverse direction of relation): 
[The Vanishing American, directed\_by, George B. Seitz], [George B. Seitz, \textasciitilde directed\_by, The Last of the Mohicans], [George B. Seitz, \textasciitilde directed\_by, Love Finds Andy Hardy], [The Last of the Mohicans, in\_language, English], [Love Finds Andy Hardy, in\_language, French]. Then, \sag generates two final reasoning paths: (The Vanishing American-George B. Seitz-The Last of the Mohicans-English), and (The Vanishing American-George B. Seitz-Love Finds Andy Hardy-French). Finally, \sag generates an answer and returns to \oag (\textit{i.e.}, English, French in the given example).

\section{Model Spec} \label{sec:model_spec}
Please check Table~\ref{sec:llm_spec}. Entries labeled as ``Unknown'' indicate that the information is not publicly available.

\section{Discussion for the Low Metric Scores for WebQSP, CWQ, and CRONQUESTIONS} \label{sec:upperbound_reason}
Based on the result shown in Table~\ref{tab:2agent_Operator_result}, \modelname showed slightly lower performance in the micro F1 score, sample-wise F1 score, and hit rate for WebQSP and CWQ (< 90\%), and micro F1 score for CRONQUESTIONS (< 50\%). Through insights from the original paper and our empirical experiments, we observed that these datasets exhibit inherent limitations. This suggests the existence of upper bounds that make achieving 100\% performance infeasible even though they are used in many previous research.

In the original WebQSP paper \cite{webqsp}, experts manually annotated 50 sample questions by constructing corresponding SQL queries. The reported correctness of these annotations ranged from 92\% to 94\%, indicating that even human-generated queries did not achieve perfect accuracy. This highlights the intrinsic difficulty of achieving complete evidence collection for every query in the WebQSP dataset. Although the original paper of CWQ does not provide a quantitative correctness analysis, the dataset is derived from WebQSP and inherits similar structural issues.

Through our own experiments, we observed several such limitations in both WebQSP and CWQ. The most critical issue was that, in some cases, a correct answer (other than the ground-truth label) entity could be retrieved using a semantically equivalent but different relation than the one used in the annotated query; however, such answers were not included in the ground-truth labels. Other minor issues included mismatches between the recorded entity and the actual answer, and inconsistencies between the question and the label (\textit{e.g.,} questions asking for "two states" but only one state being annotated for the label).

These factors collectively suggest that achieving 100\% performance in terms of F1 scores or hit rate is practically infeasible on these datasets. As a result, the slightly lower scores we observe can be attributed, at least in part, to these dataset-inherent limitations rather than model deficiencies alone.

CRONQUESTIONS exhibits high variance in the number of labels per question—some questions have a single answer, while others require many. As the number of label of the question increases, the corresponding subgraph lengthens, leading to token length issue or prediction failures. Even when the reasoning path is correct, covering all labels becomes challenging, which substantially impacts the micro F1 score. However, as seen from the sample-wise F1—where CRONQUESTIONS still achieves 89\%—\modelname generally demonstrates strong reasoning capability even under such challenging condition.

\section{\modelname Dual-Agent Approach Combined with Self-Consistency Strategy} \label{sec:dual_ensemble}
Table~\ref{tab:ablation_Supervisoragent_ensemble_result} shows the result of combining the self-consistency strategy with the dual-agent approach. The \sag generated the final answer based on three trials, leading to stricter predictions. As a result, coverage was lower compared to using dual-agent \modelname alone. For WebQSP and MetaQA, the F1 score was lower than that of a single trial of the dual-agent \modelname, whereas the hit rate was significantly higher. This is because, applying the strict self-consistency technique, some multi-label predictions were filtered out, meaning the model did not perfectly match all ground truth labels but still correctly predicted at least one. For CRONQUESTIONS, coverage, F1 scores, and hit rate were relatively lower. This dataset contains a significantly higher number of ground truth labels than others, making it difficult for any single trial to cover all labels. Consequently, the final prediction lacked sufficient labels. In FactKG, coverage varied widely, ranging from 10\% to 50\% depending on the reasoning path method. However, the hit rate consistently remained above 93\%, indicating strong performance. Overall, for multi-label tasks with many ground truth labels, a single trial using \modelname: a single trial of dual-agent approach performed more effectively than dual-agent with self-consistency strategy, suggesting that dual-agent with self-consistency strategy is not always beneficial for complex multi-label reasoning tasks.

\section{\modelname Using the Full Dataset } \label{sec:all_dataset}
Table~\ref{tab:supplement_allmini} presents the results obtained using the full dataset. In this experiment, both the \oag and \sag were set to GPT-4o mini, and the experimental setup remained identical to the main experiment.

For MetaQA 3-hop, CRONQUESTIONS, and FactKG, the hit rate exceeded 90\%, with CRONQUESTIONS reaching an impressive 98.6\%. However, coverage was generally lower or similar compared to the main experiment. This decline is likely due to the \sag{}'s limited ability to construct the correct reasoning path using the triple set during inference, as it was replaced with GPT-4o mini instead of GPT-4o. Although sufficient evidence was available, the \sag struggled to appropriately combine the necessary components of the query, leading to failed predictions. These results further highlight the critical role of the \sag in the reasoning process.

Despite the slight performance drop compared to the main experiment due to the relatively low-capacity \sag, the framework still significantly outperforms baseline methods. The effectiveness of the \textit{abstention mechanism} remains evident, ensuring that the system generates reliable predictions while maintaining robustness against uncertainty.

\section{Qualitative Analysis} \label{sec:qualitative_analysis}
\begin{figure*}
    \centering
    \includegraphics[width=\textwidth]{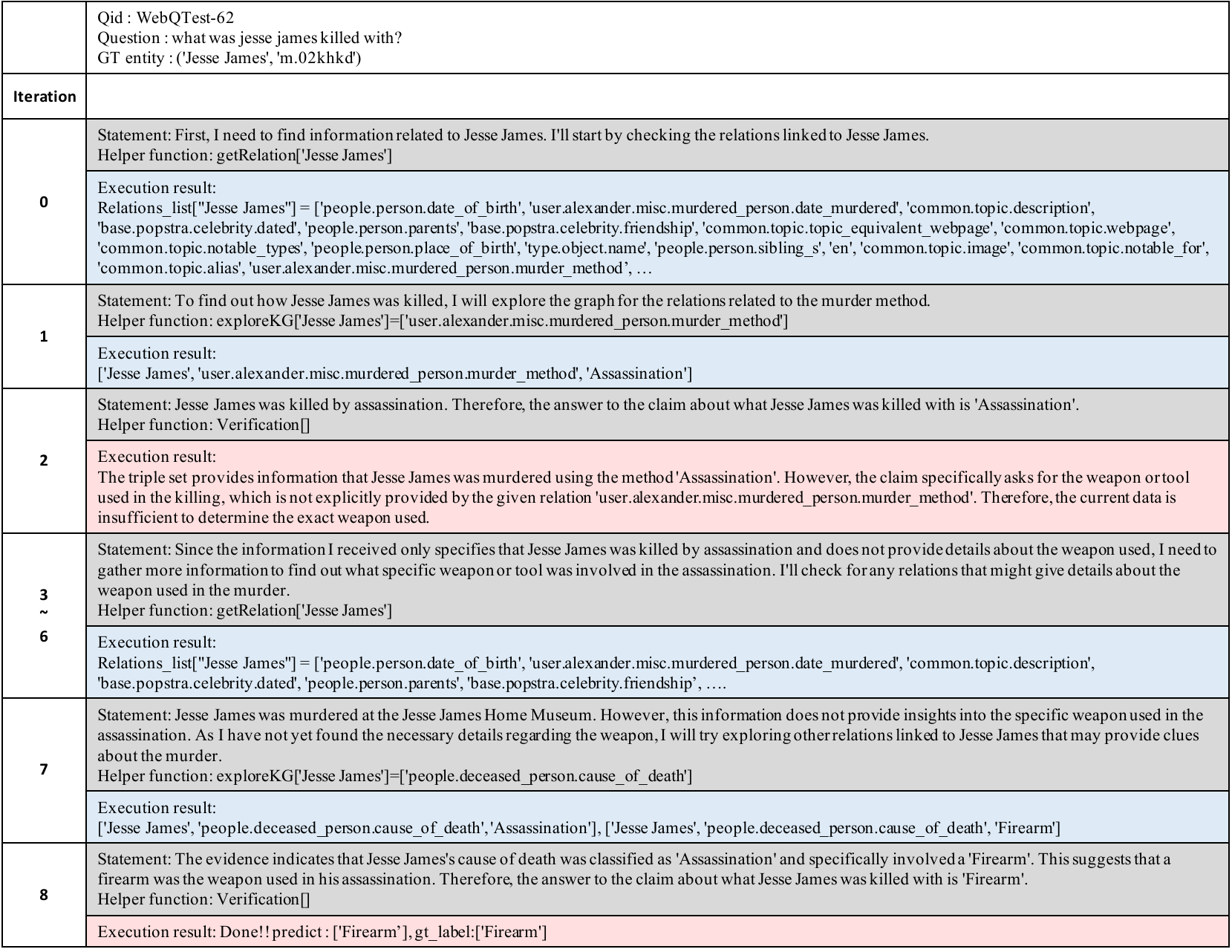}
    \caption{Successful Case in WebQSP. \sag effectively guides the model to extract a more relevant answer for the question. \textcolor{gray}{\oag}, \textcolor{blue}{\textit{Server} Response}, \textcolor{red}{\sag} for each colored box. }
    \label{fig:correct_case}
\end{figure*}

\begin{figure*}
    \centering
    \includegraphics[width=\textwidth]{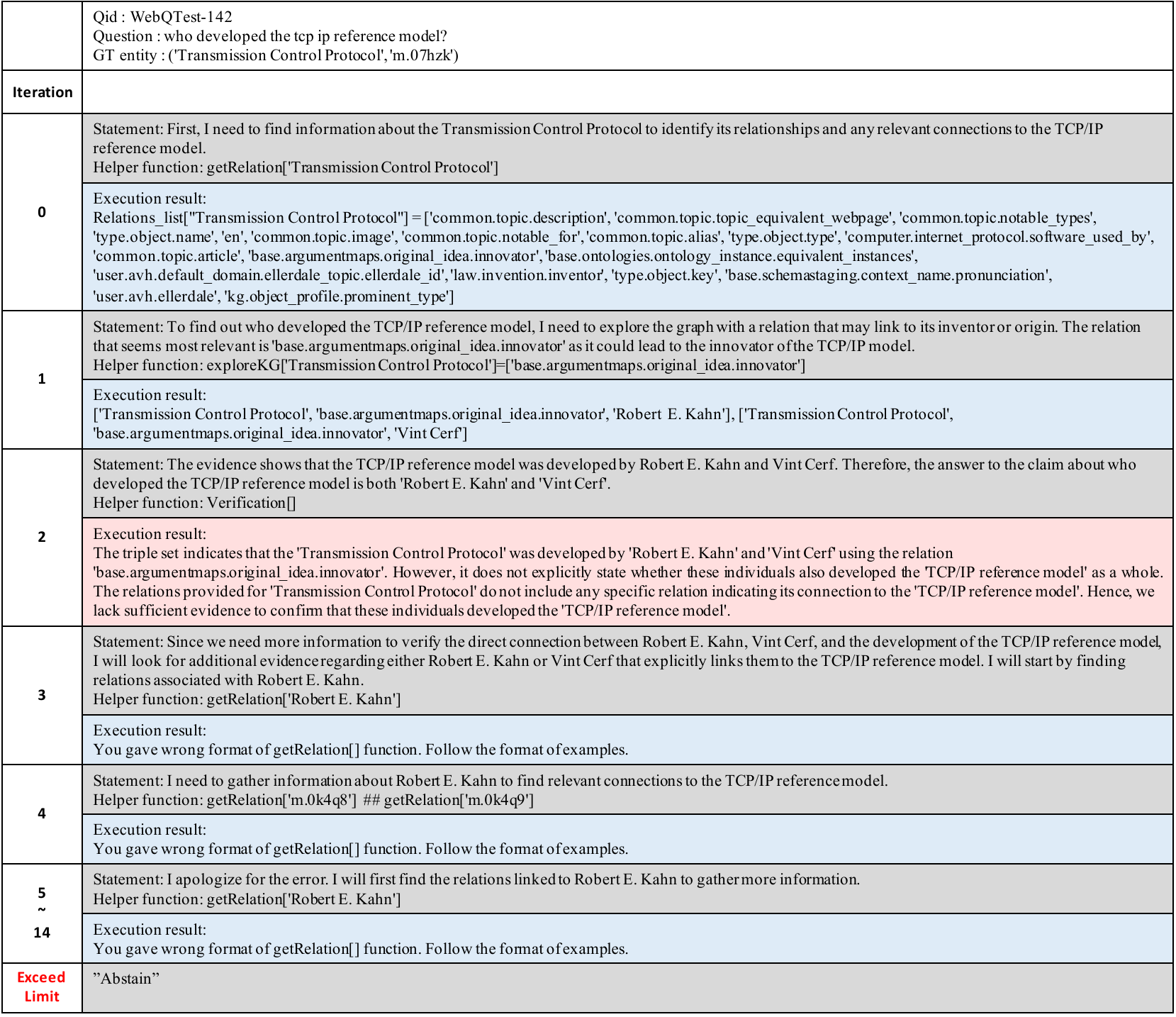}
    \caption{Failure Case in WebQSP. \sag fails to infer, leading the \oag to invoke functions in the wrong format repeatedly. \textcolor{gray}{\oag}, \textcolor{blue}{\textit{Server} Response}, \textcolor{red}{\sag} for each colored box. }
    \label{fig:wrong_case}
\end{figure*}

Figure~\ref{fig:correct_case} illustrates an example where \modelname successfully performs reasoning on WebQSP, while Figure~\ref{fig:wrong_case} shows a case where it fails. Within the \textit{T} of 15, each box represents the \oag{}'s reasoning (gray), the \textit{Server}'s execution result (blue), and the \sag{}'s reasoning (red). Some parts of the iteration process have been omitted due to excessive length.

\section{Details of Single-Agent Version of \modelname combined with Self-Consistency Strategy} \label{sec:single_agent_method}
\begin{figure*}
    \centering
    \includegraphics[width=\textwidth]{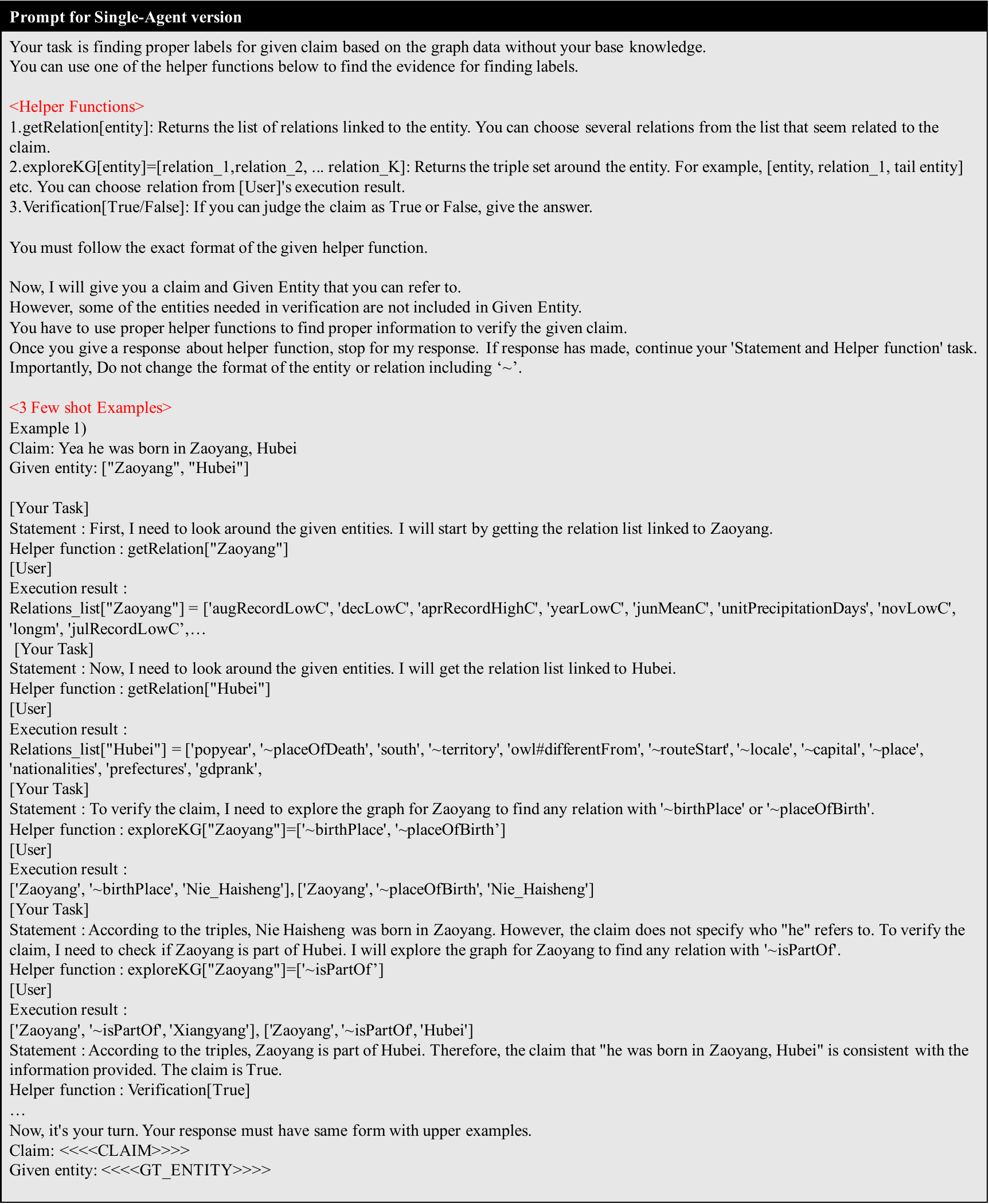}
    \caption{Prompt for single version of \modelname  [Your Task] is generated by the \oag, while [User] represents the \textit{Server}'s response.}
    \label{fig:single_prompt}
\end{figure*}

\begin{figure*}
    \centering
    \includegraphics[width=\textwidth]{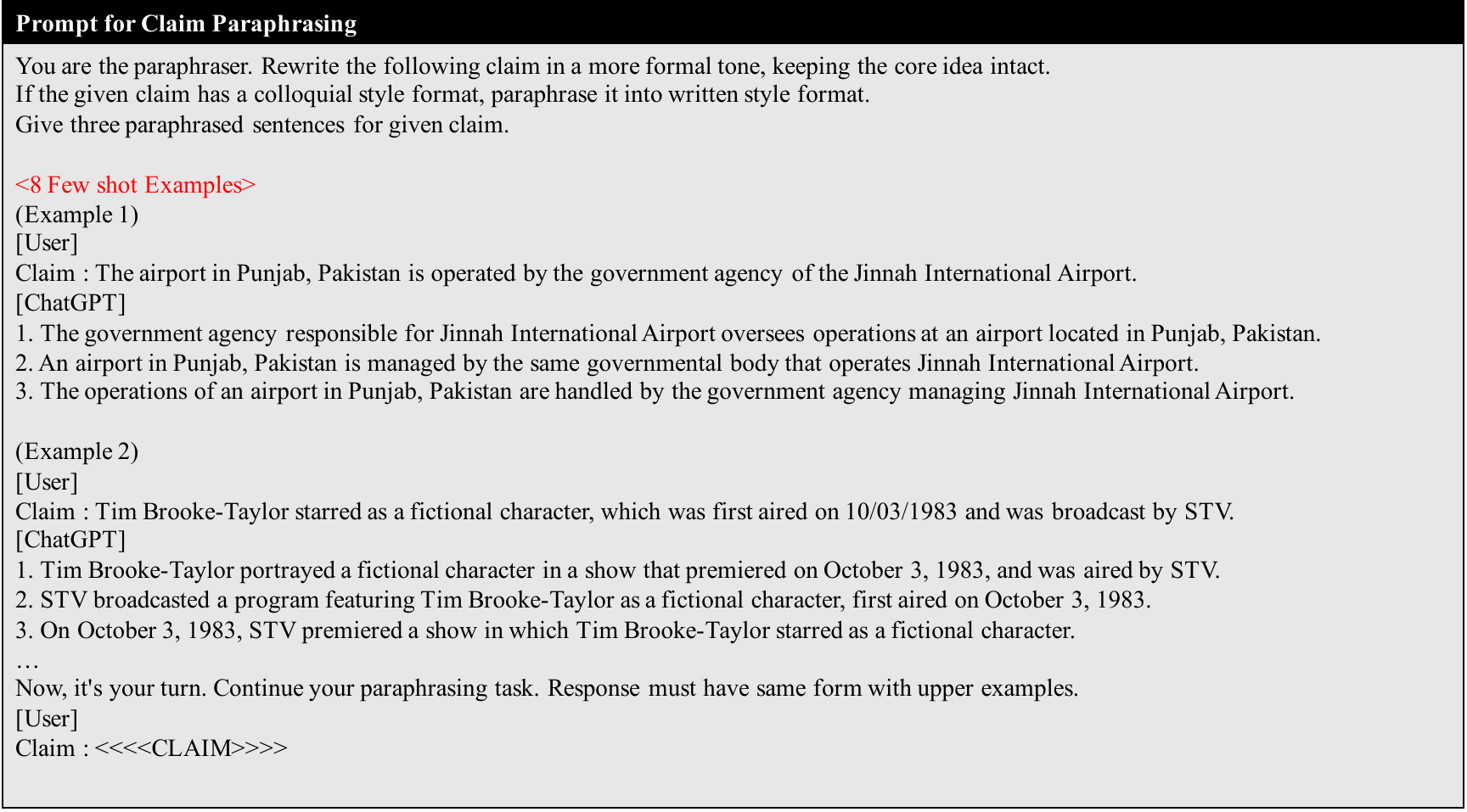}
    \caption{Prompt for query paraphrasing. [User] contains the query to be paraphrased, while [ChatGPT] generates three different variations of the sentence.}
    \label{fig:paraphrase}
\end{figure*}

The typical Self-Consistency strategy allows the language model to generate multiple reasoning paths and selects the most common answer across all trials. In contrast, our approach applies a stricter criterion, selecting the final answer only when all trials reach unanimous agreement.
The details of various reasoning paths to generate multiple responses are as follows;
The prompt used for the single-agent approach, where the \oag handles both KG retrieval and answer generation, is shown in Figure~\ref{fig:single_prompt}. For the Multi-Prompt approach, the same base prompt was used, with only the few-shot examples adjusted for in-context learning. The prompt used for query paraphrasing is identical to that in Figure~\ref{fig:paraphrase}. In this approach, each query is rewritten into three semantically equivalent but structurally different forms, and each variation is processed independently by a low-capacity LLM for three reasoning attempts. The parameter combinations used for LLM Top-p / Temperature variation are as follows:  
(Top-p, Temperature) = (0.3, 0.5), (0.7, 1.0), (0.95, 0.95)

\section{LLM Usage Statistics and Latency Analysis} \label{sec:latency}
KG-GPT requires at least 3 calls (\textit{i.e.}, Sentence Segmentation, Graph Retrieval, and Inference) to a high-capacity LLM, and ToG makes a minimum of 4 and maximum of 25 such requests, depending on the number of reasoning paths, which is closely related to the depth and width limit (\textit{i.e.}, hop limit, beam search width limit in KGs) used for ToG hyperparameter. 
This represents a substantial reduction compared to \modelname’s invocation of high‑capacity LLMs fewer than 1.43 times per query. In other words, \modelname dramatically lowers model usage costs while ensuring superior performance. One might also worry about added latency from \sag–\oag interactions; however, measuring per‑query inference times for the single‑agent and dual‑agent versions reveals only a 0.97s difference, which we confirm is not practically problematic.

\section{Token Efficiency and Cost Analysis}
In Section \ref{sec:token_usage}, we presented the token usage data, which is summarized in Table \ref{tab:cwq_token}. This section provides a detailed analysis of the associated token costs.

As of July 28, 2025, under OpenAI’s pricing policy GPT‑4o‑mini charges USD 0.15 (input) + USD 0.60 (output) per 1M tokens, and GPT‑4o charges USD 2.50 (input) + USD 10.00 (output) per 1M tokens. Therefore, on CWQ the single‑agent version (19,309 tokens with GPT‑4o) costs about USD 0.0483 and the dual‑agent version (18,987 tokens with GPT‑4o‑mini + 4,912 tokens with GPT‑4o) costs USD 0.0153 per query. In this way, by clearly separating the Operator and Supervisor roles, we achieve a USD 0.033 per query saving compared to using a high‑capacity LLM for the entire process.

\section{Experimental Setting for Baselines}\label{sec:baseline_setting}
Among the baselines used in the experiment, ToG allows width and depth to be set as hyperparameters. In our experiments, the depth was set to 3 for all datasets except FactKG, where it was set to 4. By default, ToG's width is set to 3, meaning it considers up to three entities or relations per step, regardless of the type of subject. However, this setting was highly ineffective for multi-label tasks. To improve its performance, we separately configured (relation-width, entity-width) to optimize results. The values used in the main experiment were as follows: FactKG, MetaQA, and WebQSP were set to (3, 7), (2, 5), and (3, 3), respectively.

Additionally, when ToG fails to retrieve supporting evidence from the KG, it generates answers based on the LLM’s internal knowledge. To ensure a fair comparison based solely on KG-derived information, we treated cases where ToG relied on internal knowledge after KG retrieval as \textit{Abstentions}. Similarly, while KG-GPT does not have a built-in \textit{abstention mechanism}, we considered instances where the model failed to generate a final answer due to errors during its three-step process (sentence segmentation, graph-retrieval, and inference)—such as token length limits or parsing failures—as \textit{Abstentions}.

For both baselines, prompt tuning was conducted to align them with each dataset. Specifically, we modified the few-shot examples extracted from each dataset while keeping the default prompt structure unchanged.

\section{Experiment for Measuring Knowledge Shortcut for \modelname and Baselines}
We further carried out an experiment to determine whether the LLM answers by drawing on its pre-trained world knowledge (internal knowledge) or by relying on the subgraph it has explored. To make this distinction, we built a synthetic KG that contradicts commonly accepted facts and ran \modelname on it. We constructed the synthetic KG/queries as follows based on MetaQA as shown in Table~\ref{tab:contrad_example}.

\begin{table}[t]   
\centering
\resizebox{\linewidth}{!}{%
\begin{tabular}{|p{0.46\linewidth}|p{0.46\linewidth}|}
\hline
\textbf{Original} & \textbf{Modified (Contradictory)} \\ \hline
\texttt{[Inception, directed\_by, Christopher~Nolan]} &
\texttt{[Inception, directed\_by, Mario~Van~Peebles]} \\ \hline
\textbf{Q:} Who is the director of \textit{Inception}? &
\textbf{Q:} Who is the director of \textit{Inception}? \\ \hline
\textbf{A:} Christopher Nolan &
\textbf{A:} Mario Van Peebles \\ \hline
\end{tabular}%
}
\caption{Original vs.\ contradictory triples and QA pairs (MetaQA)}
\label{tab:contrad_example}
\end{table}

As shown in Table~\ref{tab:metaqa_synthetic}, \modelname maintains high F1 score and hit rates even when evaluated on a KG that deliberately contradicts real-world facts, along with corresponding queries. This confirms that when a subgraph is provided, the LLM bases its reasoning on the retrieved structure rather than latent, pre-trained knowledge. In contrast, ToG shows extremely low accuracy, as it relies on internal knowledge when its KG exploration fails. These results demonstrate that \modelname not only explores KGs more effectively, but also grounds its answers in retrieved evidence—substantially mitigating knowledge shortcuts.

\begin{algorithm*}[h] 
\caption{Dual-Agent R2-KG Reasoning Process}
\label{alg:reasoning}
\KwIn{Claim $c$, Given entity $e_0$, Iteration limit $T$}
\KwOut{Final reasoning result or Abstain if limit exceeded}

$E_{\text{seen}} \gets \{e_0\}$ \tcp{Set of seen entities}  
$Relations \gets \{\}$ \tcp{Dictionary for entity-relation pairs}  
$gold\_triples \gets \{\}$ \tcp{Collected triples}  
$chat\_log \gets \{\}$ \tcp{Stored interaction logs}  
$i \gets 0$ \tcp{Iteration counter}

\While{$i < T$}{
    $response \gets \text{Opeartor\_response}(c, E_{\text{seen}}, Relations, chat\_log)$ \\
    Append $response$ to $chat\_log$ \\
    
    \uIf{$response = \text{getRelation}(e)$}{
        $list\_of\_relations \gets \text{Server\_response}(e)$ \\
        $Relations[e] \gets list\_of\_relations$ \\
        Append $list\_of\_relations$ to $chat\_log$
    }
    \uElseIf{$response = \text{exploreKG}(e, rel)$}{
        $triples \gets \text{Server\_response}(e, rel)$ \\
        $gold\_triples \gets gold\_triples \cup triples$ \\
        Append $triples$ to $chat\_log$
    }
    \uElseIf{$response = \text{verification}()$}{
        $feedback \gets \text{Supervisor\_response}(gold\_triples, Relations)$ \\
        \uIf{$feedback$ is an answer}{
            result $\gets feedback$ \\
            \textbf{Break}
        }
        \Else{
            Append $feedback$ to $chat\_log$ \\
        }
    }
    $i \gets i + 1$
}

\uIf{$i \geq T$}{
    \textbf{Return} Abstain
}
\Else{
    \textbf{Return} result
}

\end{algorithm*}

\begin{figure*}
    \centering
    \includegraphics[width=\textwidth]{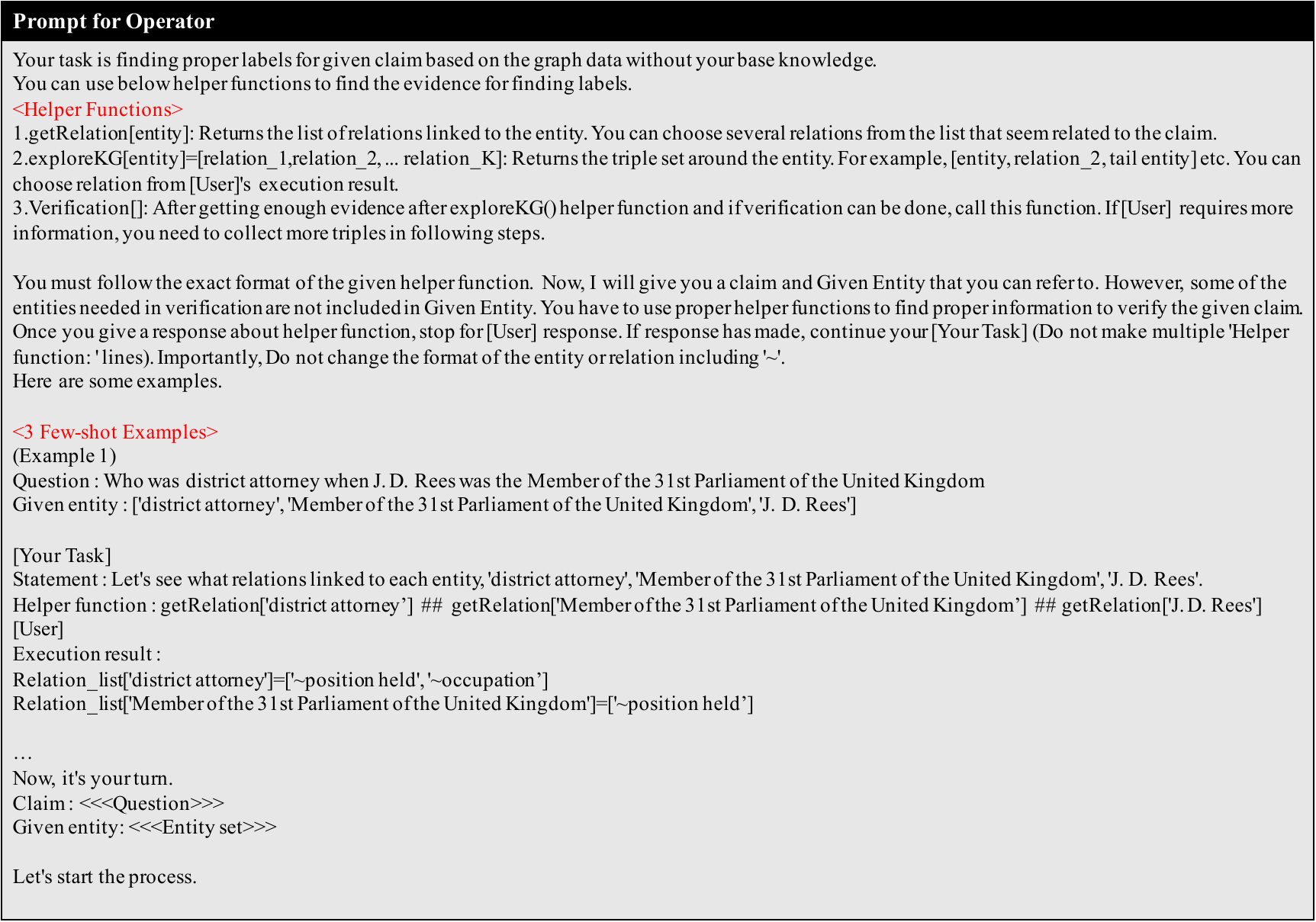}
    \caption{Used for FactKG. [Your Task] is generated by the \oag, while [User] represents either the \textit{Server}'s response or the \sag{}'s answer.}
    \label{fig:enter-label}
\end{figure*}

\begin{figure*}
    \centering
    \includegraphics[width=\textwidth]{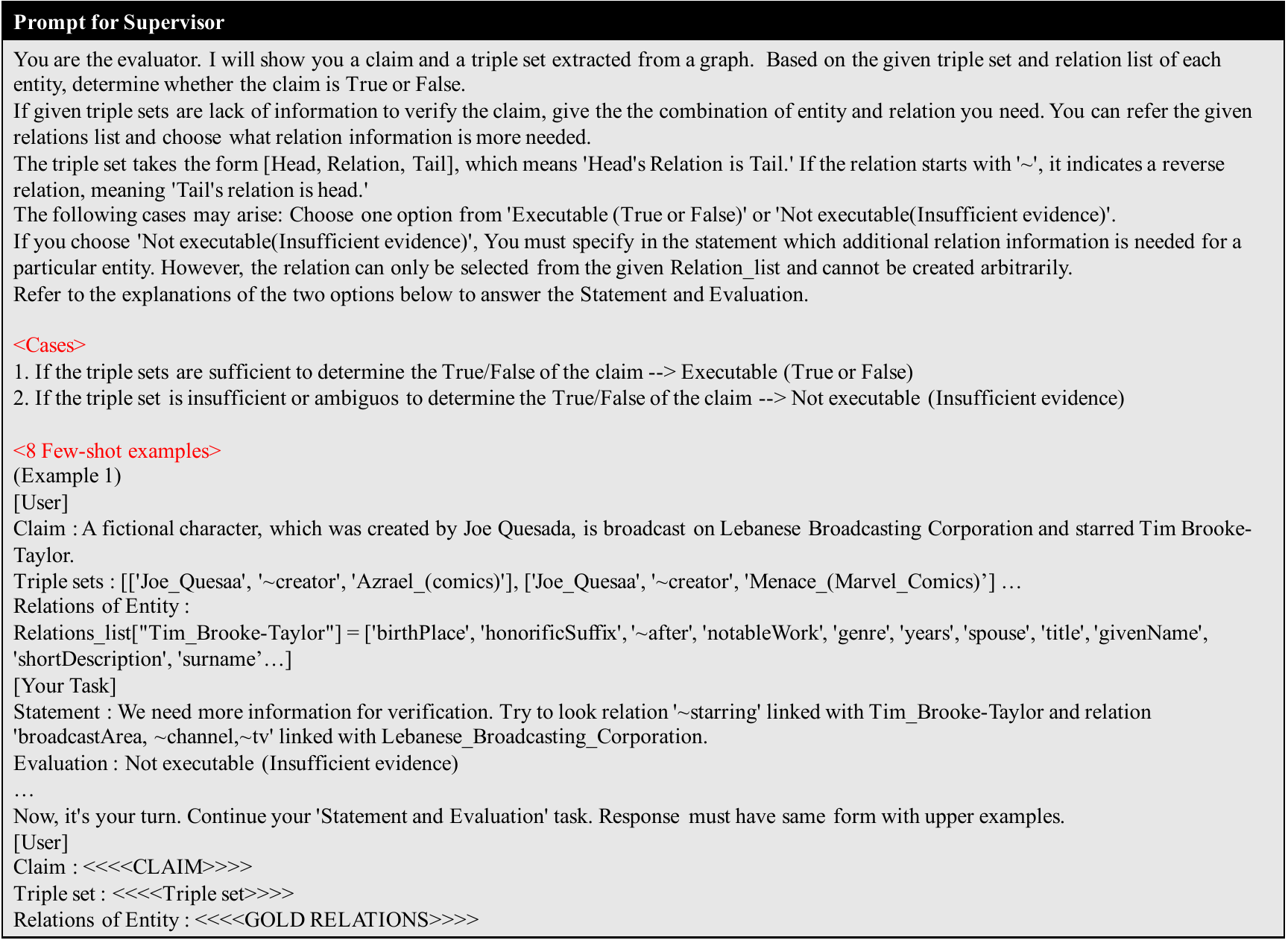}
    \caption{Used for FactKG. [Your Task] is generated by the \oag, while [User] contains the given query and the evidence collected by the \oag.}
    \label{fig:enter-label}
\end{figure*}

\end{document}